\documentclass[review]{elsarticle}

\usepackage[whole]{bxcjkjatype}
\usepackage{graphicx}

\journal{Journal of \LaTeX\ Templates}

\newcommand{\argmax}{\mathop{\rm arg~max}\limits}

\usepackage{amsmath,amssymb}
\usepackage{mathtools}
\usepackage{multirow}

\def \bftheta { {\mbox{\boldmath $\theta$}}}

\def \bfmu    { {\mbox{\boldmath $\mu$}}}

\def \bfPi     { \mathbf \Pi }
\def \bfLambda { \mathbf \Lambda }
\def \bfSigma  { \mathbf \Sigma }

\def \bfLambda { \mathbf \Lambda }

\def \rmd { \mathrm d }

\def \rmk { \mathrm k }

\def \rmK { \mathrm K }

\def \bfa { \mathbf a }

\def \bff { \mathbf f }

\def \bfs { \mathbf s }

\def \bfA { \mathbf A }
\def \bfB { \mathbf B }
\def \bfC { \mathbf C }

\def \bfI { \mathbf I }

\def \bfK { \mathbf K }

\def \bfQ { \mathbf Q }
\def \bfR { \mathbf R }
\def \bfS { \mathbf S }
\def \bfT { \mathbf T }

\def \bfW { \mathbf W }

\def \bfZ { \mathbf Z }

\def \calN { \mathcal N }
\def \calO { \mathcal O }

\usepackage[subrefformat=parens]{subcaption}
\usepackage{algpseudocode, algorithm}

\begin{document}
\begin{frontmatter}

\title{Variational Policy Search using Sparse Gaussian Process Priors for Learning Multimodal Optimal Actions}

\author[naistaddress]{Hikaru Sasaki\corref{cor1}}
\ead{sasaki.hikaru.rw3@is.naist.jp}

\author[naistaddress]{Takamitsu Matsubara}
\ead{takam-m@is.naist.jp}

\cortext[cor1]{Corresponding author}

\address[naistaddress]{Graduate School of Science and Technology, Division of Information Science, Nara Institute of Science and Technology, 8916-5 Takayamacho, Ikoma, Nara, Japan}

\begin{abstract}
Policy search reinforcement learning has been drawing much attention as a method of learning a robot control policy.
In particular, policy search using such non-parametric policies as Gaussian process regression can learn optimal actions with high-dimensional and redundant sensors as input.
However, previous methods implicitly assume that the optimal action becomes unique for each state.
This assumption can severely limit such practical applications as robot manipulations since designing a reward function that appears in only one optimal action for complex tasks is difficult.
The previous methods might have caused critical performance deterioration because the typical non-parametric policies cannot capture the optimal actions due to their unimodality. 
We propose novel approaches in non-parametric policy searches with multiple optimal actions and offer two different algorithms commonly based on a sparse Gaussian process prior and variational Bayesian inference.
The following are the key ideas: 1) multimodality for capturing multiple optimal actions and 2) mode-seeking for capturing one optimal action by ignoring the others.
First, we propose a multimodal sparse Gaussian process policy search that uses multiple overlapped GPs as a prior.
Second, we propose a mode-seeking sparse Gaussian process policy search that uses the student-t distribution for a likelihood function.
The effectiveness of those algorithms is demonstrated through applications to object manipulation tasks with multiple optimal actions in simulations. 
\end{abstract}

\begin{keyword}
Reinforcement Learning \sep Policy Search \sep Gaussian Processes \sep Multimodality \sep Mode-seeking
\MSC[2010] 00-01\sep  99-00
\end{keyword}

\end{frontmatter}


\section{Introduction}
Policy search reinforcement learning using non-parametric policy models is a promising approach for learning continuous robot control from high-dimensional, non-linear, and redundant sensory input. 
In particular, Gaussian process (or kernel) regression has successfully represented control policies in policy search \cite{KoberOP11,HoofNP17}.
Since Gaussian processes (GPs) use an implicit (high-dimensional) feature space associated with a kernel function, such high-dimensional sensor input as image features can be directly managed by a reasonable amount of training data without designing human-engineered features, unlike typical policy search methods with parametric policy models (e.g., \cite{Kober2013}) or deep neural networks that tend to require many samples \cite{Lillicrap15,Mnih16,Mnih15,GuHLL16,LevineFDA16,schulman15,Tsurumine2018}. 
For example, image-related, high-dimensional (800 dim.) features were successfully used as inputs \cite{HoofNP17}.

However, the previously proposed methods implicitly assumed that optimal actions become unique for each state.
This assumption can severely limit to such practical applications as robot manipulations since designing a reward function that appears only one optimal action for each state is often complicated for complex tasks.
A reward function that appears multiple optimal actions may lead to poor performance in the previous methods.
Multiple optimal actions may appear at particular states, although typical non-parametric policies cannot be captured due to their unimodality. 

\begin{figure*}[tb]
  \centering
  \begin{minipage}[b]{0.45\linewidth}
    \centering
    \includegraphics[width=0.55\hsize]{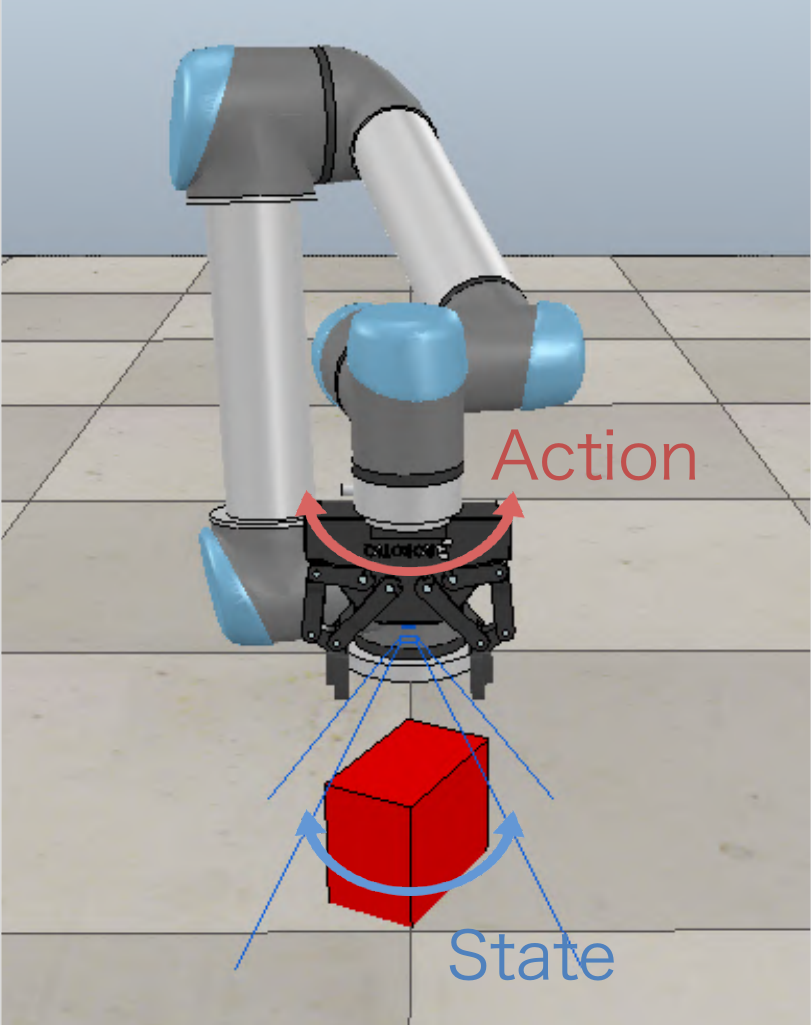} 
    \subcaption{}
    \label{fig:problem:task}
  \end{minipage}
  \begin{minipage}[b]{0.45\linewidth}
    \centering
    \includegraphics[width=0.8\hsize]{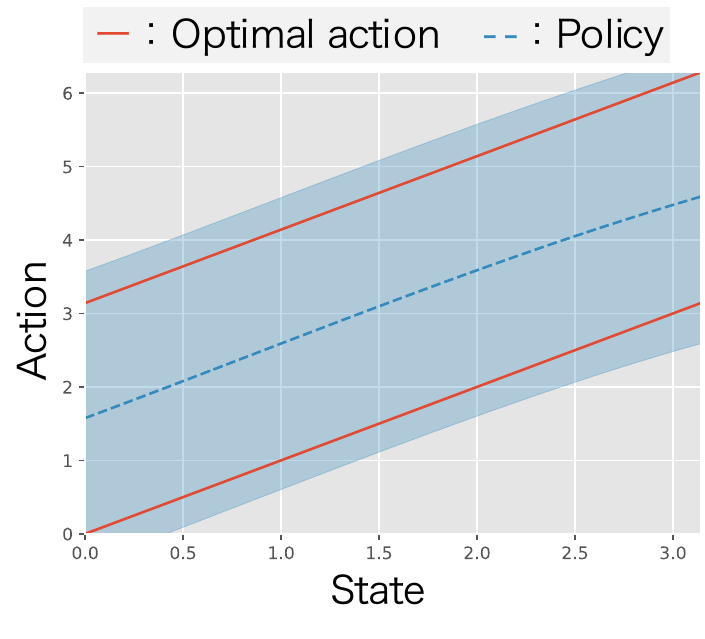} 
    \subcaption{}
    \label{fig:problem:loose}
  \end{minipage}
  \vspace{-3mm}
  \caption{Illustrative example of robotic tasks with multiple optimal actions: (a) hand-posture adjustment task environment and (b) red solid lines indicate multiple optimal actions. Blue broken line and region indicate mean and standard deviation of learned policy with a unimodal policy model, which cannot capture optimal actions.}
  \label{fig:problem}
\end{figure*}

\begin{figure*}[tb]
  \centering
  \begin{minipage}[b]{0.45\linewidth}
    \centering
    \includegraphics[width=0.8\hsize]{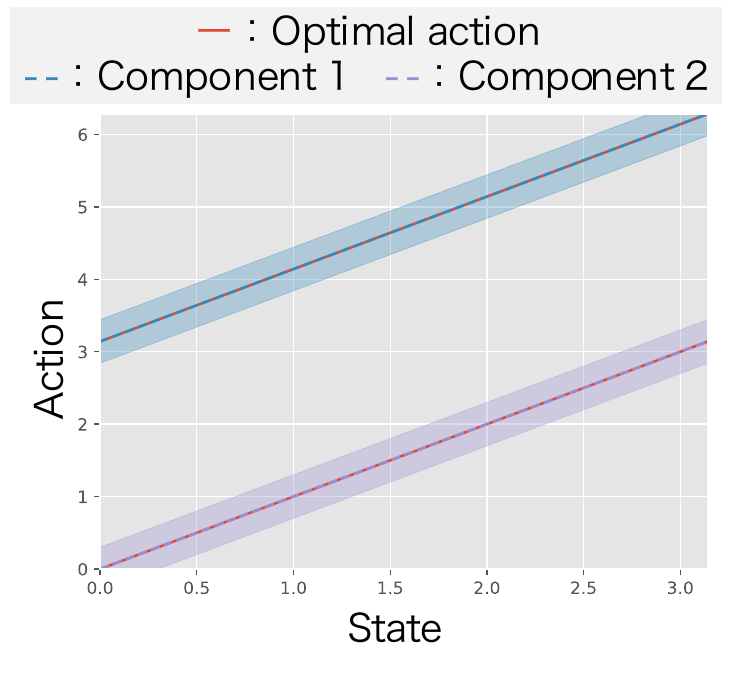}
    \vspace{-2mm}
    \subcaption{}
    \label{fig:proposed:multi}
  \end{minipage}
  \begin{minipage}[b]{0.45\linewidth}
    \centering
    \includegraphics[width=0.8\hsize]{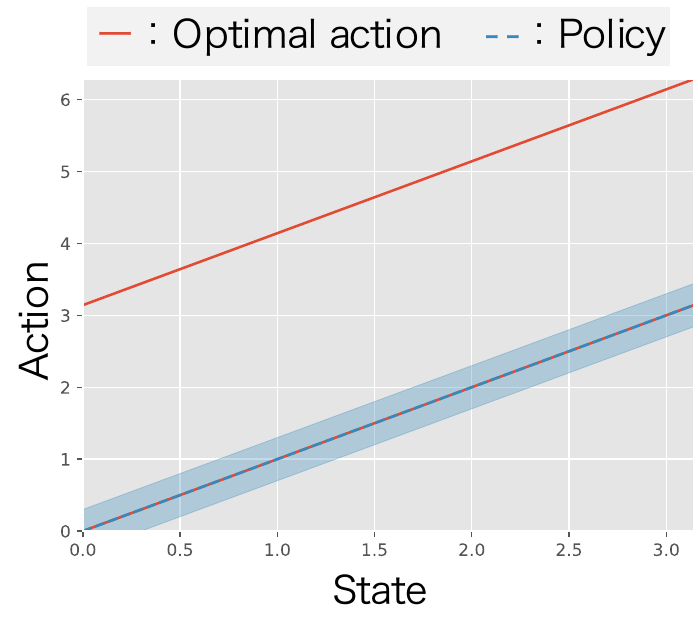} 
    \vspace{-2mm}
    \subcaption{}
    \label{fig:proposed:robust}
  \end{minipage}
  \vspace{-3mm}
  \caption{Two proposed methods for multiple optimal actions: (a) multimodal SGP-PS that can capture multimodality in optimal action using multiple models and (b) mode-seeking SGP-PS that can capture one optimal action and ignore others}
  \label{fig:proposed}
\end{figure*}

As an illustrative example of a robotic task, we consider a hand-posture adjustment task by a robotic arm (Fig. \ref{fig:problem:task}).
The objective is to grasp the object by rotating the robot's wrist.
We simply designed the reward function so that a positive value is given when the robot can grab and lift the object.
Such a reward function appears in multiple optimal actions at each state since the robot grabs the object from multiple wrist angles.
Due to the multiple optimal actions, previous methods that assume unimodality in their policy model choose incorrect actions (Fig. \ref{fig:problem:loose}).
Of course, for such simple tasks, we could elaborate the reward by adding additional terms.
However, designing a reward function that appears only one optimal action is challenging when dealing with more complex tasks. 

To alleviate this limitation, we propose novel approaches in a non-parametric policy search with multiple optimal actions and offer two different algorithms: a multimodal sparse Gaussian process policy search (multimodal SGP-PS) and a mode-seeking sparse Gaussian process policy search (mode-seeking SGP-PS).
Both methods employ sparse Gaussian processes \cite{SnelsonG05,Titsias09} as a prior of a policy model to determine a robot action and derive update laws based on variational Bayesian inference \cite{n-vipsc-11, LevineK13, luck2016sparse}.
We introduce the following key ideas: 1) multimodality for capturing multiple optimal actions (Fig. \ref{fig:proposed:multi}) and 2) mode-seeking for capturing one optimal action by ignoring the others (Fig. \ref{fig:proposed:robust}).
Multimodal SGP-PS employs a multimodal policy prior inspired by a recent extension of Gaussian processes \cite{OMGP_LAZAhROGREDILLA2012}.
This prior distribution assumes that each component is responsible for a global function over the common input space.
Mode-seeking SGP-PS employs a student-t distribution with outlier robustness as a likelihood function, which facilitates the learned policy to capture one of the optimal actions by treating the others as outliers to be ignored.
For deriving reasonable policy update schemes, we use scale mixture representation to the student-t distribution \cite{JarnoNIPS2009,Jylnki2011RobustGP}.

We validated the effectiveness of these algorithms through robotic manipulation tasks with multiple optimal actions in simulations: 1) hand-posture adjustment task using a robot simulator, V-REP \cite{V-REP} and 2) table-sweeping task using MuJoCo simulator \cite{Todorov2012}.
The results of the hand-posture adjustment task demonstrate that our methods can efficiently learn optimal actions even with multiple optimal actions that previous methods cannot. 
The multimodal SGP-PS captures multiple optimal actions with the multimodal policy model.
The mode-seeking SGP-PS learns an effective unimodal policy by seeking an optimal action.
In the table-sweeping task, we confirmed the performance of our methods for more challenging situations. 

Although a preliminary version of this paper was previously published \cite{SasakiICRA2019}, 
it only focused on an approach with multimodality for alleviating the GP-PS limitations and provided limited numerical verifications. 
Our paper significantly expanded and generalized that previous paper with a mode-seeking approach.
We thoroughly investigated its effectiveness by comprehensively analysing the experimental data, compared it with the neural-network-based methods, and discussed the remaining issues and future directions.

The remainder of this paper is structured as follows. 
Section \ref{sec:relatedwork} summarizes related works. 
Section \ref{sec:proposedmethod} presents our proposed methods, and 
Section \ref{sec:experiments} presents the simulation setup and results. 
Then several limitations and potential future works are discussed in Section \ref{sec:discussion}.
Finally, Section \ref{sec:conclusions} concludes this paper.

\section{Related Work}
\label{sec:relatedwork}

In this section, we describe related works in the following three categories: 1) non-parametric policy search, 2) multimodal reinforcement learning, and 3) robust reinforcement learning.

\subsection{Non-parametric policy search}

Non-parametric policy search uses a non-parametric model as a policy model.
Bagnell et al. \cite{Bagnell2003} embedded the function of the policy model in a reproducing kernel Hilbert space (RKHS) and used the policy gradient method to maximize the expected return, making it possible to learn non-parametric policies.
However, that approach can only learn a policy with a discrete action.
Therefore, a method was proposed for learning continuous policies by assuming a Gaussian distribution in the policy model, embedding the mean function in RKHS, and learning the policy gradient method \cite{lever2015}.
For robot applications, the Cost-regularized Kernel Regression \cite{KoberOP11} method was applied to a ball-hitting task in table tennis.
To avoid the difficulty of convergence in a gradient-based policy search, an EM-inspired non-parametric policy search was applied to a door-opening task \cite{Vien2016}.
A data efficient non-parametric policy search was proposed \cite{HoofNP17} that can learn a policy with a high-dimensional features.

Although the above studies focused on optimizing such policy models with high representation capability,
scant attention was given to the relationship between the reward function design and the limitations of unimodality in the policy.
A reward function that generates multiple optimal actions may degrade the performance in those methods.
On the other hand, we focused on this relationship and developed algorithms that can alleviate the limitations when the tackled task has multiple optimal actions.

\subsection{Multimodal reinforcement learning}

A multimodal policy search aims to learn a policy that can capture the multimodality of optimal actions.
Many previous policy search-based methods employ a hierarchical policy model that consists of low-level policies that determine robot action and a high-level policy that determines which low-level policy is used \cite{Calinon13RAS, Daniel2016, KUPCSIK2017415, OsaAAAI2018}.
However, the proposed approaches employ such hierarchical parametric models as 1) a softmax gating function with linear Gaussian sub-policies or 2) Gaussian mixture models with a parametric policy model, both of which require hand-engineered features to cope with high-dimensional sensor input, unlike non-parametric methods.

Multimodality in optimal action has also been considered in neural network-based RL methods.
For example, soft Q-learning learns a value function to acquire diverse behaviors by introducing an entropy term in the Bellman equation that promotes capturing multimodality in policies \cite{haarnoja2017}.
Soft actor-critic (SAC) is an extension of soft Q-learning; the learning performance is greatly improved \cite{haarnoja2018, haarnoja2019}.
In SAC, the value function captures various behaviors, whereas the policy employs a unimodal model.
Thus, it cannot explicitly capture multimodality in optimal actions.
SAC-GMM with a GMM policy was also explored as an early version of SAC, but its performance is worse than SAC due to the algorithmic complexity \cite{haarnoja2018D}.
Kalashnikov et al. proposed another method that selects an action from an action-value function using the cross-entropy method \cite{kalashnikov2018}.

We propose a multimodal SGP-PS that employs a policy model inspired by the overlapping mixtures of Gaussian processes (OMGP) \cite{OMGP_LAZAhROGREDILLA2012}.
Multimodal SGP-PS learns a policy whose components in the mixture are global and overlapping in state space, unlike a typical mixture model that dictates the input space for capturing non-stationarity \cite{tresp2000, rasmussen2001, meeds2006, yuan2008}.
We incorporate this feature into the policy search.
We also employ a sparse Gaussian process as a prior distribution for each low-level policy.
Such a simpler structure with a non-parametric prior allows us to derive policy update algorithms which can cope with high-dimensional sensor input without requiring hand-engineered features.

\subsection{Robust reinforcement learning}

Robust reinforcement learning acquires a robust policy for several kinds of noise.
Recent research has attracted attention by focusing on the errors between simulations and real environments in sim-to-real domains.
Reinforcement learning (robust to state noise \cite{tobinICRA2017}) and robust reinforcement learning (robust to disturbances in actual environments) have been proposed \cite{pinto17a}.
In research that focuses on reward functions, methods have been proposed that learn reward functions from human demonstrations and action evaluations \cite{fu2018learning, CHRISTIANONIPS2017} and reinforcement learning robust to noise in rewards \cite{SugiyamaICRA2009, wang2019reinforcement}.

Other methods have been proposed for capturing one optimal action even when multiple optimal actions exist.
Trust region policy optimization (TRPO) can seek optimal actions by conservatively updating policies \cite{schulman2015}.
In this method, a conservative policy update retards policy updates due to a characteristic of learning stabilization.
A deep deterministic policy gradient (DDPG) also focuses on choosing one optimal action using a deterministic policy model \cite{Lillicrap15}.
These methods are implicit mode-seeking policy search methods.
For them, it is commonly necessary to manually design many parameters in such neural network models as network structure and hyperparameters for each task.

We propose a mode-seeking SGP-PS, which effectively solves the issue of the multimodality of optimal actions.
Using student-t distribution as a likelihood, a mode-seeking SGP-PS captures one optimal action.
A non-parametric-based policy model enables efficient policy learning with few task-specific hyperparameters even with a small number of data samples.

\section{Proposed Method}
\label{sec:proposedmethod}

In this section, we derive two different non-parametric policy search algorithms: a multimodal SGP-PS and a mode-seeking SGP-PS.
Each method is derived by extending a sparse Gaussian process policy search (SGP-PS), which is a policy search that uses SGP as a policy prior.

\subsection{Sparse Gaussian process policy search (SGP-PS)}
The objective of policy search reinforcement learning is to learn the policy that maximizes the expected return by repeating the data collection using the current policy and improving the policy using the collected data under a Markov decision process.
The environment consists of state $\bfs_t$, action $a_t$, reward $r_t$, state transition probability $p(\bfs_{t+1}\mid \bfs_t, a_t)$, and initial state probability $p(\bfs_1)$.
The agent selects an action by following the current policy $\pi(a_t\mid\bfs_t)$.
After executing $T$ steps of action selection with $\pi$, the agent obtains trajectory $d = \{\bfs_1,  a_1, \cdots, \bfs_{T-1},  a_{T-1}, \bfs_T\}$, and its probability $p_\pi(d)$ is described:
\begin{align}
  p_\pi(d) &= p(\bfs_1) \prod_{t=1}^T \pi( a_t\mid \bfs_t)p(\bfs_{t+1}\mid \bfs_t, a_t).
\end{align}
Return function $R(d)$ is defined as the sum of reward $r_t$ which is obtained at time step $t$ as $R(d) = \sum_{t=1}^T r_t$.
The expected return is denoted by $J(\pi)$:
\begin{align}
 J(\pi) &= \int R(d)p_\pi(d) \mathrm d d.
\end{align}
Finally, the policy search goal is formulated:
\begin{align}
  \pi^* \leftarrow \argmax_{\pi} J(\pi).
\end{align}

We derive a policy search that optimizes the SGP policy model by maximizing a lower bound on the expected return.
We denote a policy model that determines the action from the state as $p(a_t\mid\bfs_t) = \calN(a_t\mid f(\bfs_t), \sigma^2)$, where $f$ is the mean function of the policy.
We place a GP prior on mean function $f$:
\begin{align}
  f \sim \mathcal{GP}(\mathbf 0, \mathrm k(\bfs, \bfs')),
\end{align}
where $\mathrm k(\bfs,\bfs')$ is a kernel function.

As preparation for a policy search, we reduce the computational complexity by \textit{augmenting} the prior distribution by introducing common pseudo inputs $\bar\bfS = \{\bar\bfs_l\}_{l=1}^L$ and corresponding pseudo outputs $\bar\bff=\{\bar f_l\}_{l=1}^L$ \cite{SnelsonG05, Titsias09}.
We place a GP prior on the pseudo outputs:
\begin{align}
  p(\bar\bff\mid\bar\bfS) &= \calN(\bar\bff\mid\mathbf 0, \rmK_{\bar\bfS}),
  \label{eq:prior} 
\end{align}
where $\bfK_{\bar\bfS}$ is the kernel gram matrix computed with $\bar\bfS$.
A prior distribution of $f_t = f(\bfs_t)$ is augmented:
\begin{align}
  p(f_t\mid\bar\bff, \bar\bfS, \bfs_t) &= \calN(f_t\mid\bfK_{\bfs_t,\bar\bfS}\bfK_{\bar\bfS}^{-1}\bar\bff, \lambda_t),
\end{align}
where $\lambda_t = \bfK_{\bfs_t} - \bfK_{\bfs_t,\bar\bfS}\bfK_{\bar\bfS}^{-1}\bfK_{\bar\bfS,\bfs_t}$, $\bfK_{\bar{\bfS}}=\rmk(\bar\bfS,\bar\bfS)$, $\bfK_{\bfs_t,\bar\bfS}=\rmk(\bfs_t, \bar\bfS)$.

To derive a policy search algorithm, we define the probability of the trajectory using a SGP policy model:
\begin{align}
  &p(d, \bff, \bar\bff\mid\bar\bfS) \equiv \nonumber \\
  &p(\bfs_1)p(\bar\bff\mid\bar\bfS)\prod_{t=1}^{T}p(a_t\mid f_t)p(f_t\mid\bar\bff,\bar\bfS,\bfs_t)p(\bfs_{t+1}\mid\bfs_t,a_t),
  \label{eq:pathprob}
\end{align}
where $\bff=\{f_t\}_{t=1}^T$.
For clarity, we omit the conditioning on $\bfS$ and $\bar\bfS$ in the remainder of this section.
The expected return is defined:
\begin{align}
  J(\bftheta) = \int R(d)p(d,\bff,\bar\bff)\rmd d\rmd\bff\rmd \bar\bff.
  \label{eq:objective}
\end{align}

The goal of a policy search is to find the hyperparameters of policy $\bftheta$ to maximize the expected return.
However, solving this optimization is difficult due to the analytical intractability of the integral.
We derive the lower bound of expected return $\log J_L(\bftheta)$ from Jensen's inequality by introducing variational distribution $q(d,\bff,\bar\bff)$:
\begin{align}
  &\log\frac{J(\bftheta)}{J_\mathrm{old}} \nonumber\\
  &= \log\int\frac{R(d)}{J_\mathrm{old}}\frac{q(d,\bff,\bar\bff)}{q(d,\bff,\bar\bff)} p(d,\bff,\bar\bff)\rmd d\rmd\bff\rmd \bar\bff \nonumber\\
  &\geq \int\frac{R(d)}{J_\mathrm{old}}q(d,\bff,\bar\bff)\log\frac{p(\bfa\mid\bff)p(\bff\mid\bar\bff)p(\bar\bff)}{q(d,\bff,\bar\bff)}\rmd d\rmd\bff\rmd\bar\bff \nonumber\\
  &\equiv\log J_L(\bftheta,q),
\end{align}
where $J_\mathrm{old}$ is the expected return using sampler $p_\mathrm{old}$ and $p_\mathrm{old}$ is a sampler of the training data using the previous policy.
We introduce variational distribution $q(d,\bff,\bar\bff)=p_\mathrm{old}(d)p(\bff\mid\bar\bff)p(\bar\bff)$ and apply Monte Carlo approximation regarding expectation w.r.t. $p_\mathrm{old}(d)$, and the lower bound becomes:
\begin{align}
  \log J_L(\bftheta,q) &\approx \int p(\bff\mid\bar\bff)q(\bar\bff)\log\frac{p(\tilde\bfa\mid\bff)p(\bar\bff)}{q(\bar\bff)}\rmd\bff\rmd\bar\bff,
  \label{eq:spgp:ps:lowerbound}
\end{align}
where $p(\tilde\bfa\mid\bff)$ is the following return weighted likelihood using Gaussian distribution: $p(\tilde\bfa\mid\bff) = \calN(\tilde\bfa\mid\bfW\bff, \sigma^2\bfI)$,
$\bfW$ is a weight based on the return express, $\bfW=\mathrm{diag}\left\{\sqrt{\frac{R(d^1)}{J_\mathrm{old}E}}\mathbf 1,\cdots,\sqrt{\frac{R(d^E)}{J_\mathrm{old}E}}\mathbf 1\right\}$, $d^e$ is $e$-th trajectory, and $\tilde\bfa$ is return weighted action samples, $\tilde\bfa = \bfW\bfa$.
The lower bound of the expected return (Eq. \ref{eq:spgp:ps:lowerbound}) is the lower bound of the return weighted marginal likelihood.

We derive an EM-like policy update law that maximizes the lower bound of expected return $J_L$ by alternately optimizing variational distribution $q(\bar\bff)$, parameters $\bftheta$ by variational Bayesian inference.

The action selection for new state $\bfs_*$ is determined by the predicted distribution of the SGP policy model using variational distribution $q(\bar\bff)$ and hyperparameter $\bftheta$:
\begin{align}
  p(a_*\mid\bfs_*) \approx \int p(a_*\mid f_*)p(f_*\mid\bar\bff,\bfs_*)q(\bar\bff) \rmd f_*\rmd\bar\bff.
\end{align}

\subsection{Multimodal sparse Gaussian process policy search (multimodal SGP-PS)}
Next we derive the multimodal SGP-PS based on SGP-PS.
To capture multiple optimal actions, we consider the following product of $M$ different GPs as a prior of function $f$ of the control policy:
\begin{align}
  f \sim \alpha\prod_{m=1}^M \mathcal{GP}^{(m)}(\mathbf 0, \rmk^{(m)}(\bfs, \bfs')).
\end{align}
We assume a multimodality control policy that uses $M$ GPs and expresses the lower bound of the return weighted likelihood product of the $M$ lower bounds:
\begin{align}
  &\log J_L(\bftheta_M, q) \approx \nonumber \\
  &\prod_{m=1}^M\int p(\bff^{(m)}\mid\bar\bff^{(m)})q(\bar\bff^{(m)})\log\frac{p(\tilde\bfa\mid\bff^{(m)})p(\bar\bff^{(m)})}{q(\bar\bff^{(m)})}\rmd\bff^{(m)}\rmd\bar\bff^{(m)}, 
  \label{eq:m:lowerbound}
\end{align}
where $\bar\bff^{(m)}$ and $\bff^{(m)}$ indicate the pseudo outputs and the function outputs of the $m$-th SGP.
Optimizing the above lower bound w.r.t. policy parameters is, however, insufficient to search for multimodal behaviors; all the $M$ GPs tend to converge to the same solution since all the training data are commonly shared to train all the GPs.
To allow exploration of different solutions for all the GPs, we introduce another latent variable, the so-called binary indicator matrix $\bfZ$ whose $nm$ element is a binary variable that is associated with $m$-th GP \cite{OMGP_LAZAhROGREDILLA2012}.
Each row in $\bfZ$ has one non-zero entry.
We assume the prior on binary indicator matrix $\bfZ$:
\begin{align}
  p(\bfZ) = \prod_{n=1,m=1}^{N,M}[\bfPi]_{nm}^{[\bfZ]_{nm}},
\end{align}
where $\bfPi$ is the $N\times M$ probability matrix and $[\bfPi]_{nm}$ indicates the probability that $a_n$ is generated by the $m$-th GP.
Moreover, we extend the return weighted likelihood function:
\begin{align}
  p(\tilde\bfa\mid\{\bff^{(m)}\}) &= \int p(\bfZ)p(\tilde\bfa\mid\{\bff^{(m)}\},\bfZ)\rmd\bfZ, \\
  p(\tilde\bfa\mid\{\bff^{(m)}\},\bfZ) &= \prod_{n=1,m=1}^{N,M}\calN(\tilde\bfa_n\mid\bfW_{nn}\bff, \sigma^2)^{[\bfZ]_{nm}}.
\end{align}
Due to the analytical intractability of the integral of $\bfZ$, we apply Jensen's inequality:
\begin{align}
  &\log J'_L(\bftheta_M, q) = \nonumber\\
  &\int p(\{\bff^{(m)}\}\mid\{\bar\bff^{(m)}\})q(\{\bar\bff^{(m)}\})q(\bfZ) \cdot\nonumber\\
  &\log\frac{p(\tilde\bfa\mid\{\bff^{(m)}\},\bfZ)p(\{\bar\bff^{(m)}\})p(\bfZ)}{q(\{\bar\bff^{(m)}\})q(\bfZ)}\rmd\{\bff^{(m)}\}\rmd\{\bar\bff^{(m)}\}\rmd\bfZ, 
\end{align}
where $q(\{\bar\bff^{(m)}\}) = \prod_{m=1}^M q(\bar\bff^{(m)})$, $p(\{\bar\bff^{(m)}\}) = \prod_{m=1}^M p(\bar\bff^{(m)})$ and $p(\{\bff^{(m)}\}\mid\{\bar\bff^{(m)}\}) = \prod_{m=1}^M p(\bff^{(m)}\mid\bar\bff^{(m)})$.

Finally, we derive an EM-like iterative optimization scheme.
We repeat the E-step that finds optimal variational distribution $q(\{\bar\bff^{(m)}\})$ and $q(\bfZ)$ and the M-step that finds optimal parameter $\bftheta_M$ by alternatively maximizing $\log J'_L$.
We describe the derivation of the analytical update laws of the variational distributions and the analytical form of the objective function in \ref{appendix:multi}.
A summary of the multimodal SGP-PS is given in Algorithm \ref{alg:multi}.
Table \ref{table:comp:amount:multi} shows the computational complexity of each optimization, and the complexity of $q(\{\bar\bff^{(m)} \})$ and $\log J_L'$ is reduced due to the pseudo inputs.
Although the computational complexity of $q(\bfZ)$ is increased, the overall complexity is reduced.

\begin{algorithm}[tb]
  \caption{Multimodal SGP-PS algorithm}
  \label{alg:multi}
  \begin{algorithmic}[1]
    \State Initialize $\bftheta_M$, $\bar\bfS$
    \While{reward not converged}
    \State{\# Sample $E$ trajectories}
    \For{$e=1:E$}
    \State Sample trajectory $d$ using policy $\pi$
    \EndFor
    \State{\# Policy Improvement}
    \While{$\log J_L'$ is not converged} 
    \State{\# E step}
    \While{$\log J_L'$ is not converged} 
    \State Update $q(\{\bar\bff^{(m)}\})$ with Eq. (\ref{eq:multi:update:qf})
    \State Update $q(\bfZ)$ with Eq. (\ref{eq:multi:update:qz})
    \EndWhile
    \State{\# M step}
    \State{$\bftheta_M \gets \argmax_{\bftheta_M} ~ \log J_L'$}
    \EndWhile
    \EndWhile
  \end{algorithmic}
\end{algorithm}

\begin{table}[tb]
  \centering
  \caption{Computational complexity of each optimization in multimodal SGP-PS: $N$, $M$, and $L$ are amounts of training data, components, and pseudo inputs.}
  \vspace{-2mm}
  \label{table:comp:amount:multi}
  \begin{tabular}{|c|ccc|}\hline
                      & $q(\{\bar\bff^{(m)}\})$ & $q(\bfZ)$ & $\log J_L'$   \\ \hline \hline
    Multimodal SGP-PS & $\calO(MNL^2)$          & $\calO(MNL^2)$ & $\calO(MNL^2)$ \\ \hline
  \end{tabular}
\end{table}

We analytically obtain the $M$ predictive distribution using variational distribution $q(\{\bar\bff^{(m)}\})$ and parameter $\bftheta_M$ variational learning.
The $m$-th predictive distribution of action $a^{(m)}_*$ corresponds to new state $\bfs_*$:
\begin{align}
  p(a_*^{(m)}\mid\bfs_*) \approx \int p( a_*^{(m)}\mid f_*)p(f_*\mid\bar\bff^{(m)},\bfs_*) q(\bar\bff^{(m)}) \rmd f_*\rmd \bar\bff^{(m)}. 
\end{align}
The analytical solution of the predictive distribution of the multimodal SGP policy search is described in \ref{appendix:multi:predictive}.

We employ a softmax function that uses the negative variance of the predictive distribution:
\begin{align}
  p(m) = \frac{\exp(-\sigma_*^{(m)}/\beta)}{\sum_{m=1}^M \exp(-\sigma_*^{(m)}/\beta)},
\end{align}
where $p(m)$ is the probability of using the $m$-th predictive distribution and $\beta$ is a temperature parameter that controls the trade-off between exploration and exploitation.

\subsection{Mode-seeking sparse Gaussian process policy search (mode-seeking SGP-PS)}
In this section, we derive a mode-seeking sparse Gaussian process policy search based on SGP-PS.
Mode-seeking means that the policy can capture an optimal action at each state by ignoring other optimal actions.
To capture one optimal action, the mode-seeking SGP-PS employs student-t distribution as a likelihood and estimates the reliability of each data to ignore low-reliability data.
The following is the probabilistic density function (PDF) of the student-t distribution:
\begin{align}
  &\mathrm{St}(x\mid\mu, a, b) = \nonumber\\
  &\frac{b^a}{\Gamma(a)\sqrt{2\pi}}\left[b+\frac{(x-\mu)^2}{2}\right]^{-a-1/2}\Gamma(a+1/2),
\end{align}
where $\Gamma(\cdot)$ is a gamma function.
Although the PDF is complicated, it can be described more simply by a scale mixture representation that uses Gaussian and gamma distributions:
\begin{align}
  \mathrm{St}(x\mid\mu, a, b) &= \int_0^\infty \calN(x\mid\mu,\tau^{-1})\mathrm{Gam}(\tau\mid a, b) \rmd\tau, \\
  \mathrm{Gam}(\tau\mid a,b) &= \frac{b^a}{\Gamma(a)}\tau^{a-1}\exp(-b\tau),
\end{align}
where $\tau$ is the precision of the Gaussian distribution and indicates the data reliability.
We use gamma distribution for the precision $\tau$ of the Gaussian distribution.

To learn a policy that captures one optimal action, we use the student-t distribution as the likelihood:
\begin{align}
  p(\tilde\bfa\mid\bff) &= \mathrm{St}(\tilde\bfa\mid\bff) \nonumber \\
  &= \int p(\tilde\bfa\mid\bff,\bfT)p(\bfT)\rmd\bfT,  \\
  p(\tilde\bfa\mid\bff, \bfT) &= \prod_{n=1}^N \calN(\tilde\bfa_n\mid\bff_n,\tau_n^{-1}), \\
  p(\bfT) &= \prod_{n=1}^N \mathrm{Gam}(\tau_n\mid a, b),
\end{align}
where $\bfT=\mathrm{diag}\{\tau_1,\cdots,\tau_N\}$.
The integration w.r.t $\bff$ in Eq. (\ref{eq:spgp:ps:lowerbound}) cannot be solved analytically since the likelihood is complicated.
By introducing a scale mixture representation and variational distribution $q(\bfT)$ of the data reliability, we derive a new lower bound for the mode-seeking SGP-PS:
\begin{align}
  &\log J_L'(\bftheta_R, q) = \nonumber \\
  &\int p(\bff\mid\bar\bff)q(\bar\bff)q(\bfT)\log\frac{p(\tilde\bfa\mid\bff,\bfT)p(\bfT)p(\bar\bff)}{q(\bar\bff)q(\bfT)}\rmd\bff\rmd\bar\bff\rmd\bfT.
\end{align}

We assume that $q(\bfT) = \prod_{n=1}^N q(\tau_n)$.
We derive the update laws of $q(\bar\bff)$ and $q(\tau_n)$ which maximize the lower bound of the expected return based on variational Bayesian inference.
The analytical solution of the update laws and the lower bound are described in \ref{appendix:robust}.
Alternately updating the variational distribution and the hyperparameters alternately, the mode-seeking SGP policy is learned.
This method's algorithm is described in Alg. \ref{alg:robust}.
Table \ref{table:comp:amount:robust} shows the computational complexity of each optimization, and the complexity of $q(\{\bar\bff^{(m)} \})$ and $\log J_L'$ is reduced by the pseudo inputs.
Although the computation amount of $q(\bfT)$ increased, the overall complexity was reduced.

We analytically obtained the predictive distribution of the mode-seeking SGP policy for new state $\bfs_*$ using variational distribution $q(\bar\bff)$, $q(\bfT)$, and parameters $\bftheta_R$, which are obtained in the learning:
\begin{align}
  p(a_*\mid \bfs_*) \approx \int p(a_*\mid f_*,\tau)p(f_*\mid\bar\bff,\bfs_*)q(\bar\bff)q(\tau)\rmd f_*\rmd\bar\bff\rmd\tau. 
\end{align}

The analytical solution of the predictive distribution of mode-seeking SGP-PS is described in \ref{appendix:robust:predictive}.

\begin{algorithm}[!t]
  \caption{Mode-seeking SGP-PS algorithm}
  \label{alg:robust}
  \begin{algorithmic}[1]
    \State Initialize $\bftheta_R$, $\bar\bfS$
    \While{reward not converged}
    \State{\# Sample $E$ trajectories}
    \For{$e=1:E$}
    \State Sample trajectory $d$ using policy $\pi$
    \EndFor
    \State{\# Policy Improvement}
    \While{$\log J_L'$ is not converged}
    \State{\# E step}
    \While{$\log J_L'$ is not converged}
    \State Update $q(\bar\bff)$ with Eq. (\ref{eq:robust:update:qf})
    \State Update $q(\bfT)$ with Eq. (\ref{eq:robust:update:qtau})
    \EndWhile
    \State{\# M step}
    \State{$\bftheta_R \gets \argmax_{\bftheta_R} ~ \log J_L'$}
    \EndWhile
    \EndWhile
  \end{algorithmic}
\end{algorithm}
\begin{table}[tb]
  \centering
  \caption{Computational complexity of each optimization in mode-seeking SGP-PS: $N$ and $L$ are amounts of training data and pseudo inputs. }
  \label{table:comp:amount:robust}
  \vspace{-2mm}
  \begin{tabular}{|c|ccc|}\hline
                        & $q(\bar\bff^{(m)})$ & $q(\bfT)$ & $\log J_L'$   \\ \hline\hline
    Mode-seeking SGP-PS & $\calO(NL^2)$ & $\calO(N^2)$ & $\calO(N^2)$ \\ \hline
  \end{tabular}
\end{table}

\section{Experiments}
\label{sec:experiments}
We conducted experiments with two robot control tasks in simulations: hand-posture adjustment task and table-sweeping task.
We investigated the effectiveness of our proposed methods in the hand-posture adjustment task and demonstrated scalability for more challenging situations in the table-sweeping task.

\begin{figure}[t]
  \centering
  \includegraphics[width=7cm]{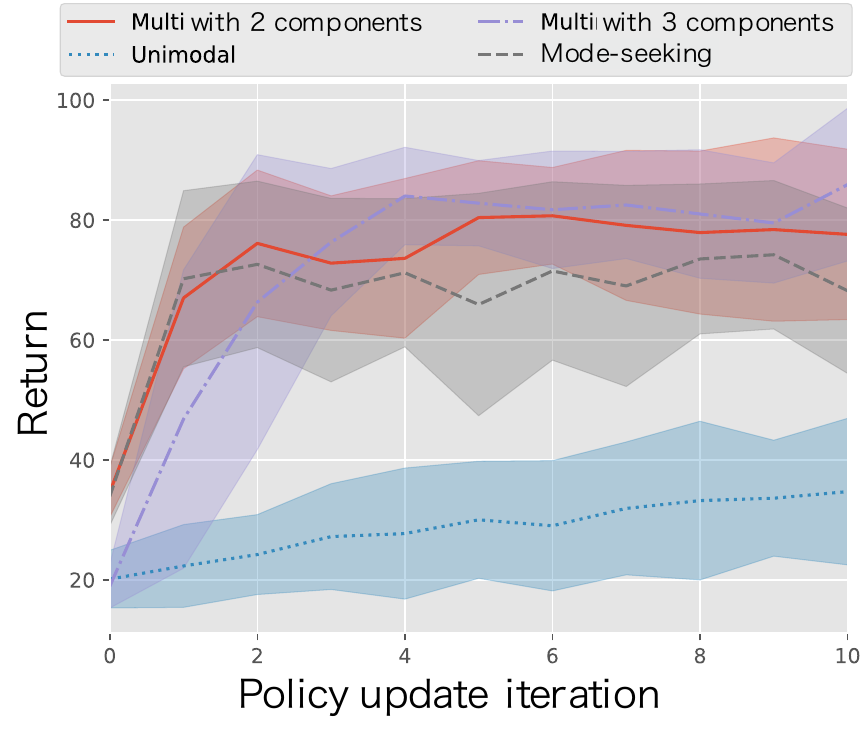}
  \vspace{-3mm}
  \caption{Learning performance of policy search for hand-posture adjustment task over ten experiments: Mean value and standard deviations are shown.}
  \label{fig:grasping:reward}
\end{figure}
\begin{figure}[t]
  \centering
  \includegraphics[width=7cm]{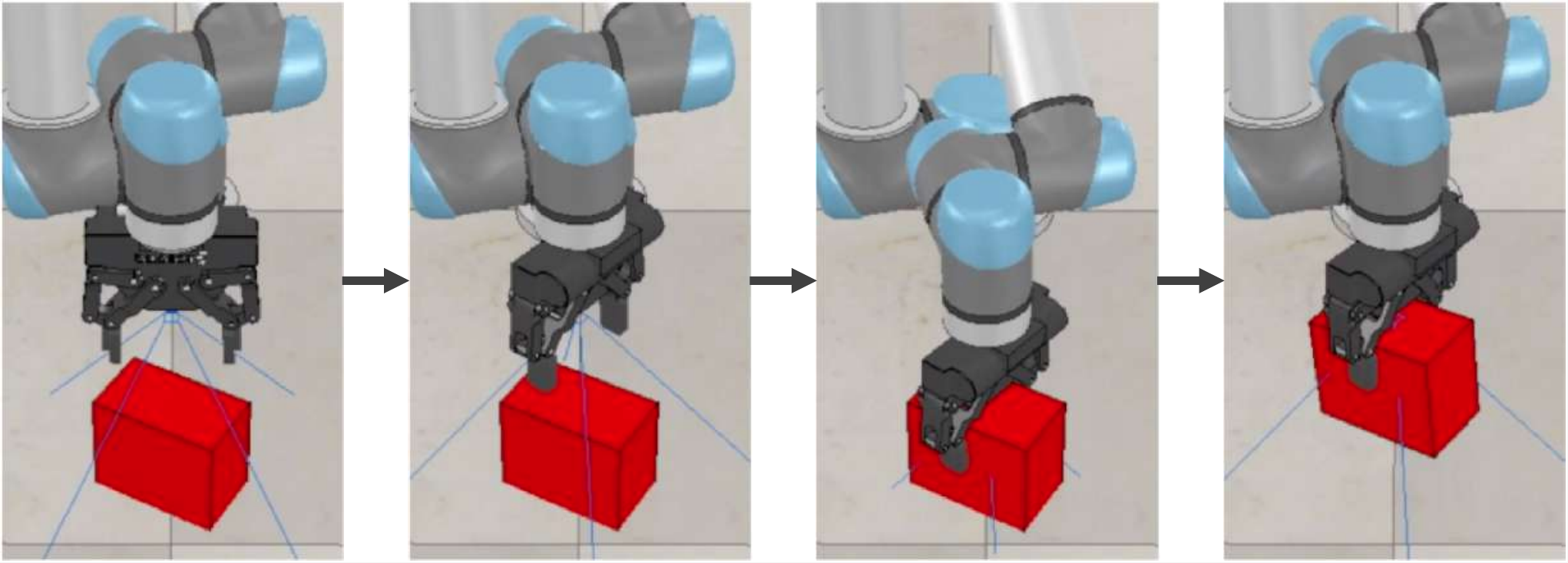}
  \vspace{-2mm}
  \caption{Robot behavior controlled with a policy learned by multimodal SGP-PS in hand-posture adjustment task}
  \label{fig:grasping:action}
\end{figure}

\begin{figure}[!t]
  \begin{minipage}[b]{\linewidth}
    \centering
    \includegraphics[width=6cm]{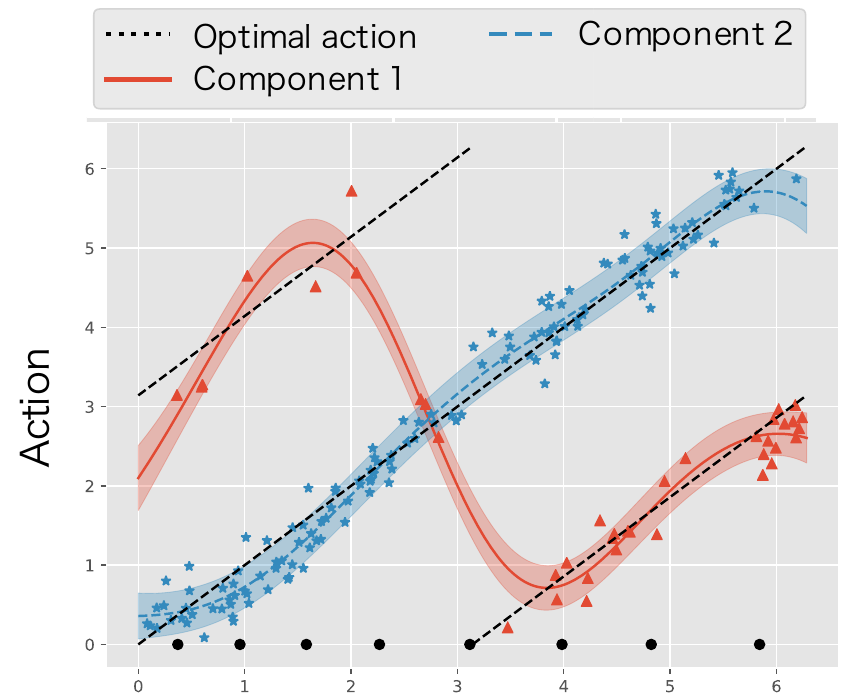}
    \vspace{-2mm}
    \subcaption{}
    \label{fig:grasping:policy:multi2}
  \end{minipage}
  \begin{minipage}[b]{\linewidth}
    \centering
    \includegraphics[width=6cm]{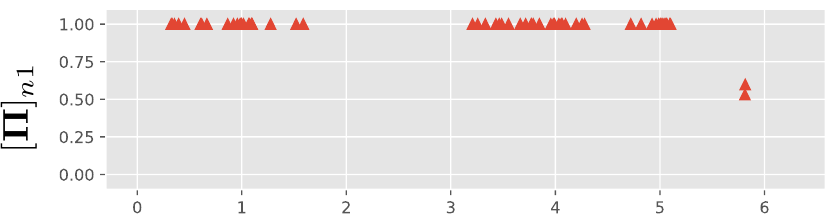}
    \vspace{-2mm}
    \subcaption{}
    \label{fig:grasping:policy:multi2:pi1}
  \end{minipage}
  \begin{minipage}[b]{\linewidth}
    \centering
    \includegraphics[width=6cm]{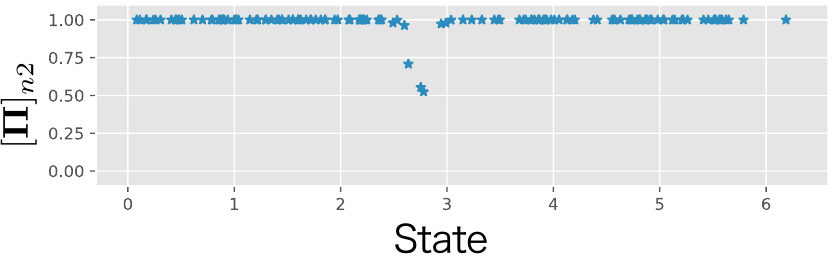}
    \vspace{-2mm}
    \subcaption{}
    \label{fig:grasping:policy:multi2:pi2}
  \end{minipage}
  \vspace{-6mm}
  \caption{(a) Learned policy for hand-posture adjustment task by multimodal SGP-PS with two components. Black dots indicate pseudo input. (b) and (c) show posterior probability of the data association of each data point shown in (a).}
  \label{fig:grasping:multi2}
\end{figure}

\begin{figure}[!t]
  \begin{minipage}[b]{\linewidth}
    \centering
    \includegraphics[width=6cm]{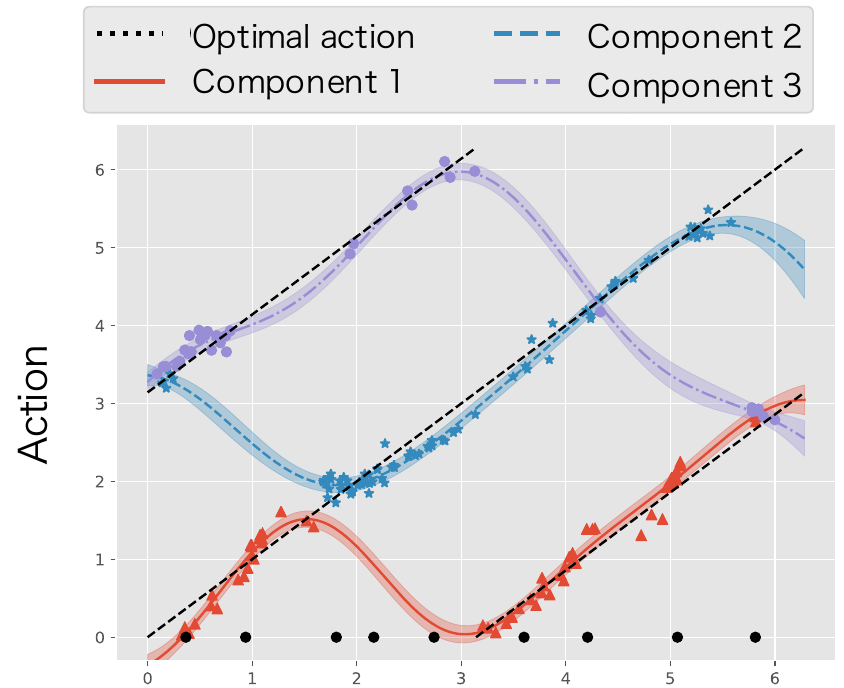}
    \vspace{-2mm}
    \subcaption{}
    \label{fig:grasping:policy:multi3}
  \end{minipage}
  \begin{minipage}[b]{\linewidth}
    \centering
    \includegraphics[width=6cm]{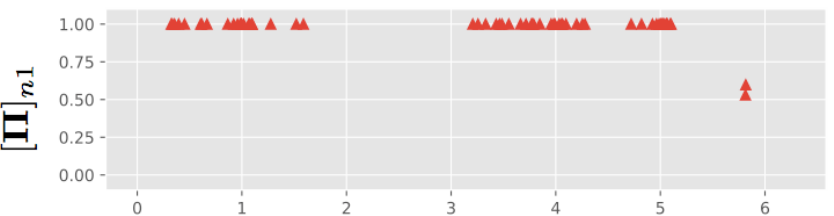}
    \vspace{-2mm}
    \subcaption{}
    \label{fig:grasping:policy:multi3:pi1}
  \end{minipage}
  \begin{minipage}[b]{\linewidth}
    \centering
    \includegraphics[width=6cm]{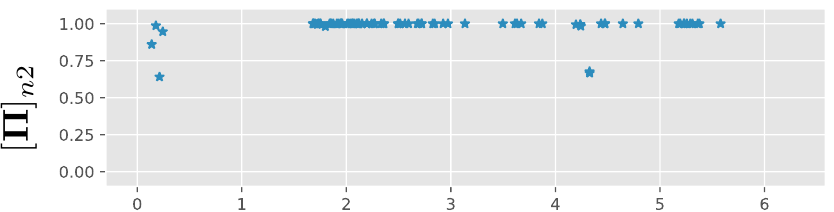}
    \vspace{-2mm}
    \subcaption{}
    \label{fig:grasping:policy:multi3:pi2}
  \end{minipage}
  \begin{minipage}[b]{\linewidth}
    \centering
    \includegraphics[width=6cm]{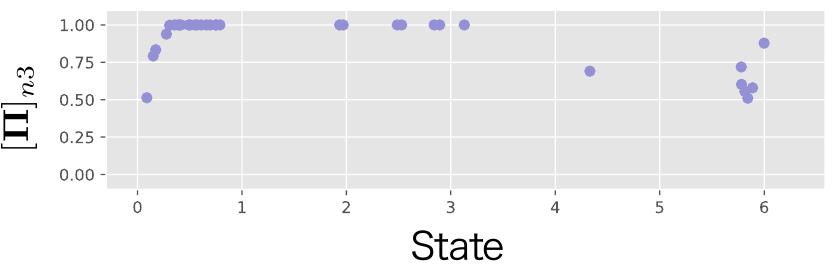}
    \vspace{-2mm}
    \subcaption{}
    \label{fig:grasping:policy:multi3:pi3}
  \end{minipage}
  \vspace{-6mm}
  \caption{(a) Learned policy for hand-posture adjustment task by multimodal SGP-PS with three components. Black dots indicate pseudo input. (b), (c), and (d) show posterior probability of data association of each data point in (a).}
  \label{fig:grasping:multi3}
\end{figure}

\begin{figure}[t]
  \begin{minipage}[b]{\linewidth}
    \centering
    \includegraphics[width=6cm]{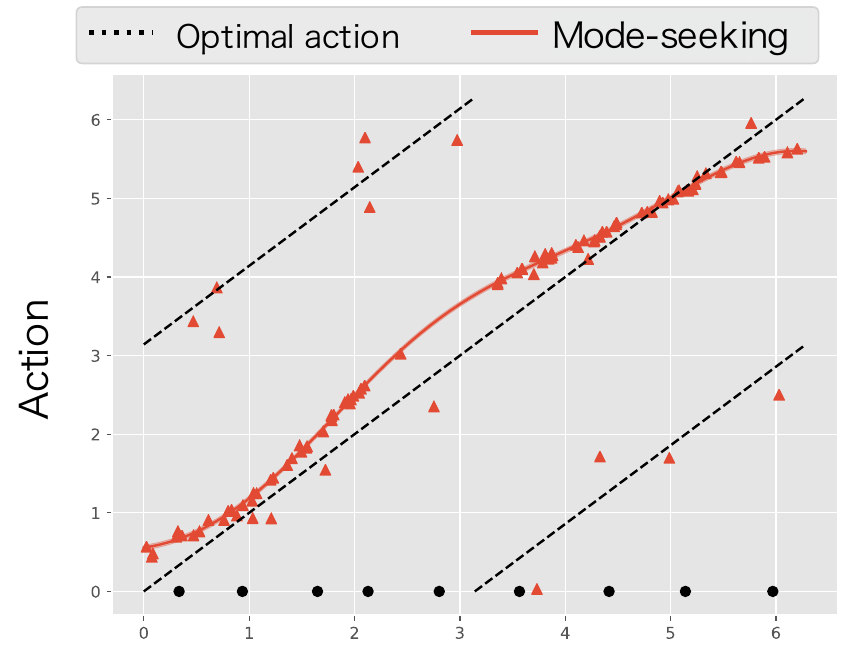}
    \vspace{-2mm}
    \subcaption{}
    \label{fig:grasping:policy:robust}
  \end{minipage}
  \begin{minipage}[b]{\linewidth}
    \centering
    \includegraphics[width=6cm]{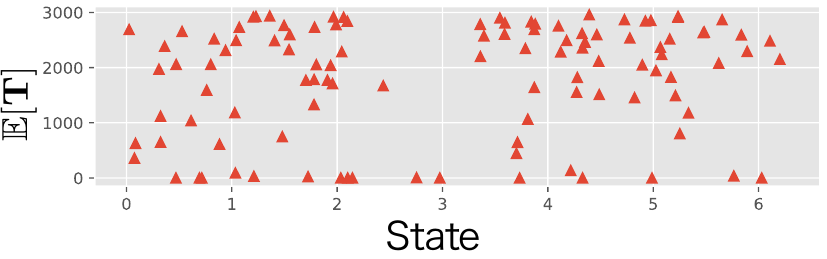}
    \vspace{-2mm}
    \subcaption{}
    \label{fig:grasping:policy:robust:tau}
  \end{minipage}
  \vspace{-6mm}
  \caption{(a) Learned policy for hand-posture adjustment task by mode-seeking SGP-PS. Black dots indicate pseudo input. (b) posterior probability of data association.}
  \label{fig:grasping:robust}
\end{figure}

\subsection{Hand-posture adjustment task}
\label{section:exp1:grasp}

\subsubsection{Settings}
We experimented with the simple robot task shown in Fig. \ref{fig:problem:task} to confirm the performance of our proposed methods on the task with the multiple optimal actions.
The task's environment consisted of a UR5 robot arm and a red, 7$\times$12$\times$10 cm$^3$ cube.
The robot arm has a two-fingered gripper and can rotate its wrist in one revolution.
The task aims to grasp the red cube by its longer side; due to its physical constraint, the robot cannot grasp its shorter side.
The robot can grasp the object in two ways: by rotating its wrist left or right.
So, the task has two optimal actions.

The state is the orientation of the red cube.
The action is the rotation angle of the robot wrist between $-\pi$ rad to $\pi$ rad.
After determining the action, the robot moves to grasp the object with its wrist angle decided as an action.
Since each episode is terminated in one step, the length of the trajectory is $T=1$.
Return function $R(d)$ is binary.
$R(d)=100$ if the robot grasps the cube, otherwise $R(d)=0$.

We confirm the performances of our methods to the multiple optimal actions by comparing them with a unimodal SGP-PS that employs a standard SGP as a policy model. 
We set the number of components of the multimodal SGP-PS to two and three to confirm its performance with different numbers of components to the number of multiple optimal actions.
We used an isotropic-squared exponential kernel in each method.
In this experiment, each method explored $E = 100$ episodes and learned a policy using the data of 100 episodes and the best 80 episodes as sample reuse after exploration.
The number of pseudo-inputs of each method was set: $L=20$.

\subsubsection{Results}
Figure \ref{fig:grasping:reward} shows the mean and the standard deviation of the return over ten experiments using multimodal SGP-PS with two and three components, mode-seeking SGP-PS, and unimodal SGP-PS.
The policies learned by our proposed methods indicate higher performance than that obtained by unimodal SGP-PS.
This result suggests that the expansions of a policy model in our proposed methods are valid for the task with the multiple optimal actions.
Moreover, multimodal SGP-PS can learn an appropriate policy even when the number of components is redundant.
Fig. \ref{fig:grasping:action} shows the robot behavior using a learned multimodal SGP policy which, can select an appropriate action for the object's angle.
Fig. \ref{fig:grasping:policy:multi2} shows a learned policy by multimodal SGP-PS with two components.
The learned multimodal SGP policy can captures multiple optimal actions indicated by the black dotted line by properly estimating data associations.
To capture two optimal actions, component 1 captures two dotted lines, and component 2 captures the middle line.
Figs. \ref{fig:grasping:policy:multi2:pi1} and \ref{fig:grasping:policy:multi2:pi2} show the optimized posterior distribution of $\bfZ$, which is the data points assigned to each component.
In the crossing point of the two components, the posterior of the data association indicates low probability.
Fig. \ref{fig:grasping:policy:multi3} shows a learned policy by multimodal SGP-PS with three components. 
Each component of the learned policy captures one of three dotted lines.
Figs. \ref{fig:grasping:policy:multi3:pi1}, \ref{fig:grasping:policy:multi3:pi2}, and \ref{fig:grasping:policy:multi3:pi3} show the optimized posterior distribution of $\bfZ$.
Similar to the case of the multimodal SGP-PS with two components, the posterior of the data association indicates a low probability of the two components' crossing points.
Fig. \ref{fig:grasping:policy:robust} shows a learned policy by mode-seeking SGP-PS.
The policy acquired by it captured one of the optimal actions at each state, coincidentally in a similar way as with two components in Fig. \ref{fig:grasping:policy:multi2}.
Fig. \ref{fig:grasping:policy:robust:tau} shows the posterior distribution of reliability $\bfT$.
The mode-seeking SGP-PS learned a unimodal policy by estimating the reliability of the data and ignoring low-reliability data.

In summary, the simulation results in the hand-posture adjustment task suggest the effectiveness of our proposed methods for learning a policy in tasks with multiple optimal actions.

\subsection{Table-sweeping task}
\subsubsection{Settings}
\begin{figure}[!t]
  \centering
  \begin{minipage}[b]{3.5cm}
    \centering
    \includegraphics[width=0.8\hsize]{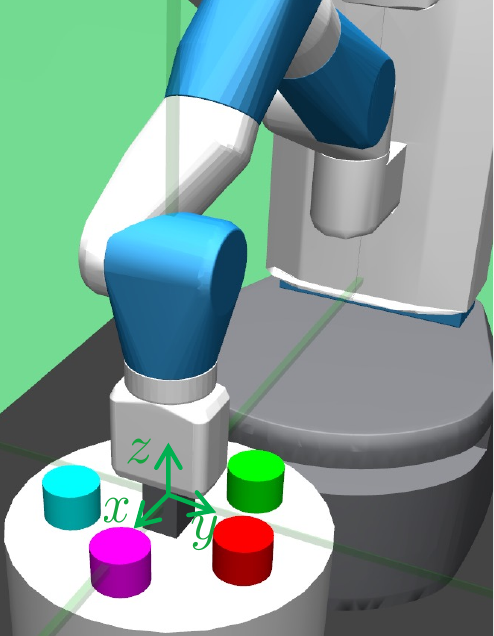}
    \vspace{-1mm}
    \subcaption{}
    \label{fig:fetch:sweep:env:overview}
  \end{minipage}
  \begin{minipage}[b]{3.5cm}
    \centering
    \includegraphics[width=0.8\hsize]{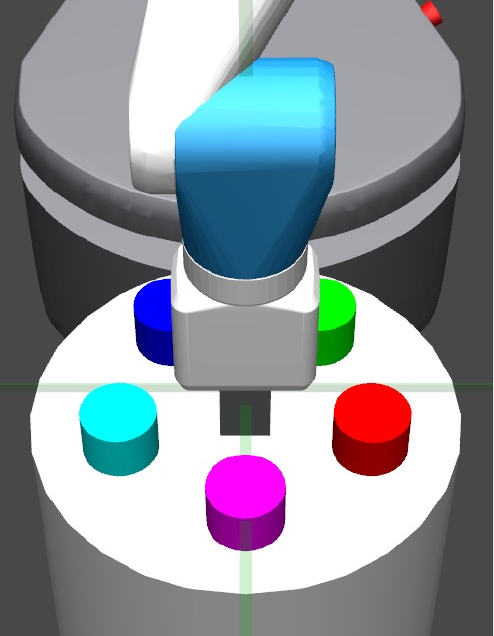}
    \vspace{-1mm}
    \subcaption{}
    \label{fig:fetch:sweep:env1}
  \end{minipage}
  \begin{minipage}[b]{3.5cm}
    \centering
    \includegraphics[width=0.8\hsize]{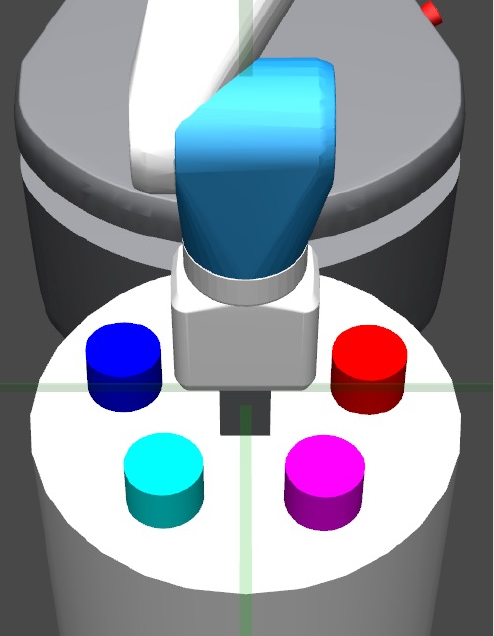}
    \vspace{-1mm}
    \subcaption{}
    \label{fig:fetch:sweep:env2}
  \end{minipage}
  \vspace{-3mm}
  \caption{Table-sweeping task environment with five objects. (a) is a overview of table-sweeping task. Green vectors show axes of the Cartesian coordinate to represent a position of the end-effector and objects. (b) and (c) indicate two different initial positions of five objects.}
  \label{fig:fetch:sweep:env}
\end{figure}

The aim of the table-sweeping task is to have the fetch robot sweep the five objects on the table individually by robot's end-effector in the environment shown in Fig. \ref{fig:fetch:sweep:env}.
In this task, we learned an individual action policy based on the number of objects by exploiting the fact that this task can be naturally decomposed into subtasks according to the number of objects.
Thus, five individual policies are learned.
We constructed the table-sweeping task environment based on the fetch environment proposed in \cite{plappert2018} and implemented it as a learning environment in OpenAI Gym.
The table-sweeping task is more challenging than the hand-posture adjustment task for the following reasons:
\begin{enumerate}
\item The number of optimal actions is up to five.
\item Multiple steps are required to accomplish the task.
\item The state space has a higher dimension.
\end{enumerate}

The task environment consists of the fetch robotic arm, a 40-cm diameter table, and five cylindrical objects with diameters of 7 cm, and heights of 2.5 cm.
State consists of the two-dimensional Cartesian position of the end-effector and five objects, and the two-dimensional Cartesian relative position of the five objects related to the position of the end-effector.
Each axis of the Cartesian coordinate is shown in Fig. \ref{fig:fetch:sweep:env:overview}.
Therefore, a state is represented by a 22-dimensional vector consisting of positions of the end-effector and five objects and relative positions of five objects.
The definition of a state is relied on \cite{plappert2018}.
We set that the position and the relative position in a state corresponding to swept objects are assigned 0.
An action is a two-dimensional vector and each dimension specifies the desired end-effector movement in each axis of two-dimensional Cartesian coordinates on the table.
The absolute value of each dimension of action is limited to 4 cm or less so that even an optimal policy takes multiple steps to achieve the task.
Return function $R(d)$ rewards the trajectory data with 10 when an object is swept and punishes $0.1\times T$ for the length of the episode $T$.

We set that the end-effector's position is initialized to the center of the table after sweeping an object.
The positions of the five objects are initialized by uniformly sampling from two patterns shown in Fig. \ref{fig:fetch:sweep:env}.
The five objects are arranged at 72-degree intervals on a circle with a diameter of 12 cm around the z-axis, but there is 32-degree difference between two patterns.
The task is terminated when an object is swept or when the length of the trajectory has reached to $T=20$.
After the robot swept an object, the end-effector moves to the center of the table to execute the next task.

To compare our methods, we employed unimodal SGP-PS, SAC \cite{haarnoja2018}, and TRPO \cite{schulman2015}.
SAC and TRPO are state-of-the-art reinforcement learning methods that learn neural-network policies for continuous action space.
The hyperparameters of SAC and TRPO are described in Tables \ref{table:sac:params} and \ref{table:trpo:params}.

We used a squared exponential kernel with automatic relevance determination in the SGP-based methods.
The number of pseudo-inputs of each method is set to $L=20$.
We set the number of objects on the table as the number of components of the multimodal SGP-PS.
For a fair comparison, the SGP-based methods and the NN-based methods as TRPO and SAC commonly update their policies using every 1000 steps worth of exploration episodes.

\begin{figure*}[!t]
  \centering
  \begin{minipage}[b]{0.32\linewidth}
    \centering
    \includegraphics[width=1.0\hsize]{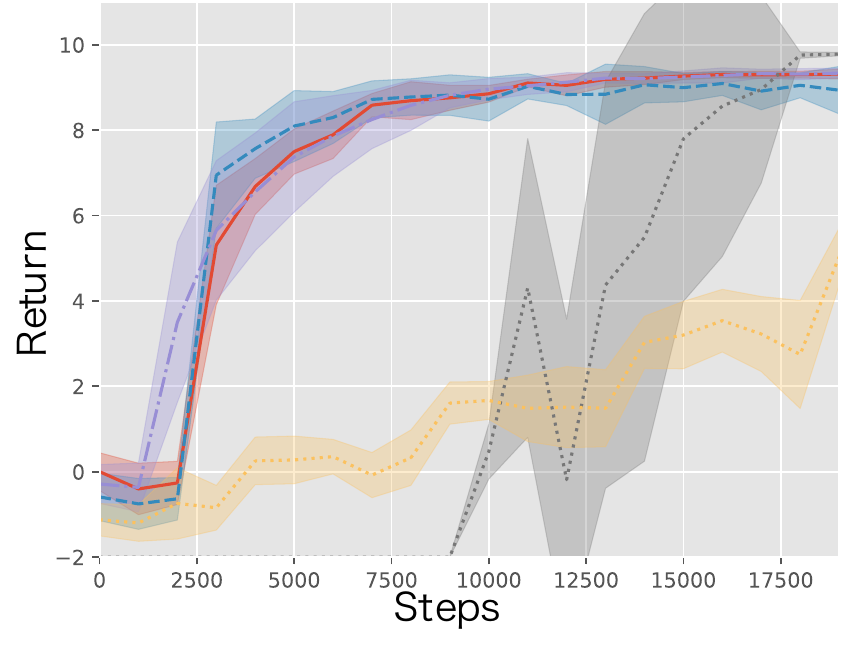}
    \vspace{-8mm}
    \subcaption{One object}
    \label{fig:sweep1}
  \end{minipage}
  \begin{minipage}[b]{0.32\linewidth}
    \centering
    \includegraphics[width=1.0\hsize]{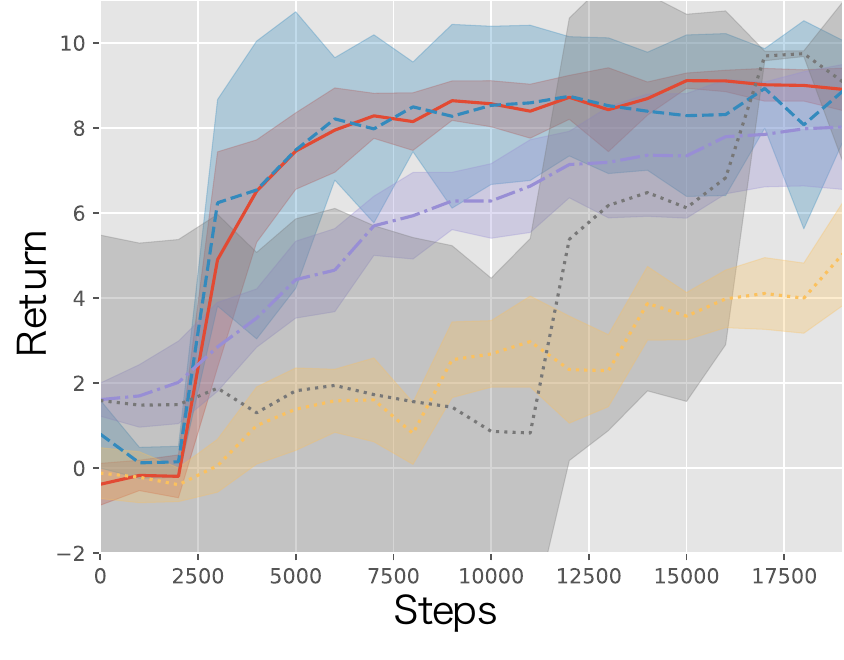}
    \vspace{-8mm}
    \subcaption{Two objects}
    \label{fig:sweep2}
  \end{minipage}
  \begin{minipage}[b]{0.32\linewidth}
    \centering
    \includegraphics[width=1.0\hsize]{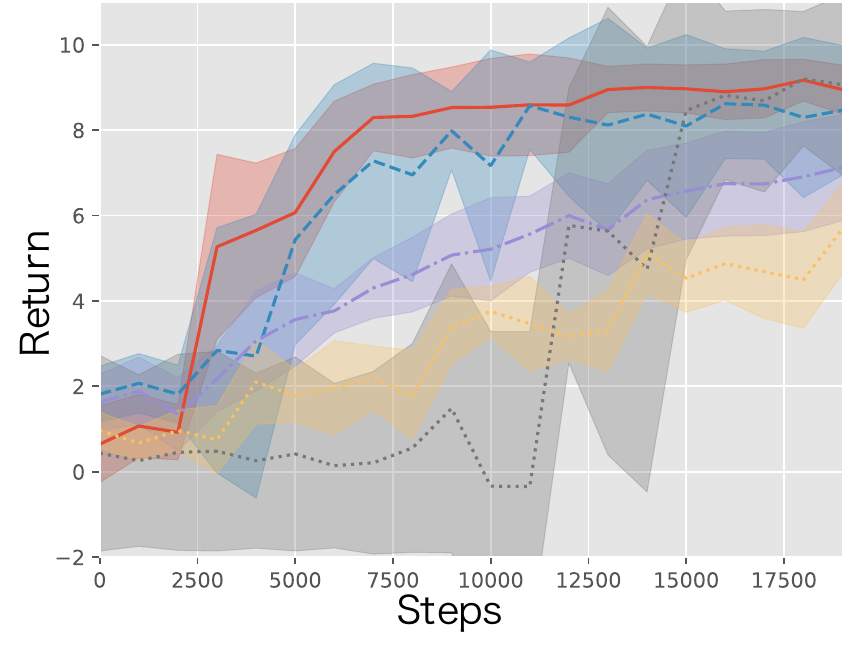}
    \vspace{-8mm}
    \subcaption{Three objects}
    \label{fig:sweep3}
  \end{minipage}\\
  \vspace{2mm}
  \begin{minipage}[b]{0.32\linewidth}
    \centering
    \includegraphics[width=1.0\hsize]{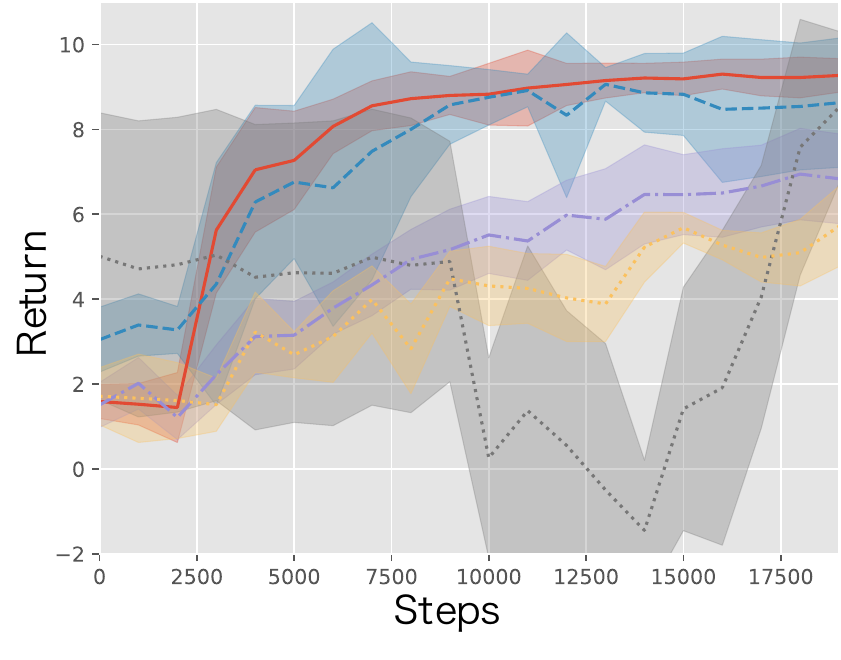}
    \vspace{-8mm}
    \subcaption{Four objects}
    \label{fig:sweep4}
  \end{minipage}
  \begin{minipage}[b]{0.32\linewidth}
    \centering
    \includegraphics[width=1.0\hsize]{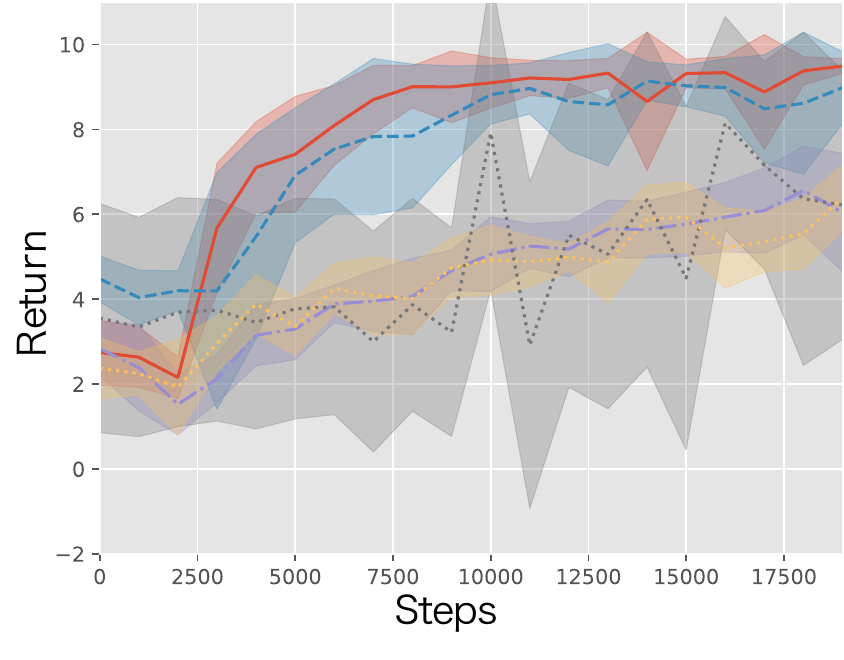}
    \vspace{-8mm}
    \subcaption{Five objects}
    \label{fig:sweep5}
  \end{minipage}
  \begin{minipage}[b]{0.32\linewidth}
    \centering
    \includegraphics[width=0.4\hsize]{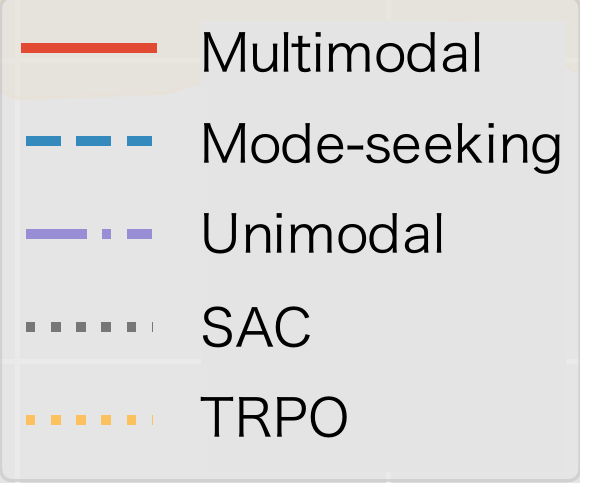}
    \vspace{2mm}
    \subcaption{Legends of each figure}
    \label{fig:sweep:legend}
  \end{minipage}
  \vspace{-1mm}
  \caption{Learning performances of multimodal SGP-PS, mode-seeking SGP-PS, unimodal SGP-PS, SAC, TRPO in table-sweeping task with one to five objects: Each curve is averaged over ten experiments.}
  \label{fig:fetch:sweep:result}
\end{figure*}

\subsubsection{Results}
\begin{figure*}
  \centering
  \begin{minipage}[b]{0.3\linewidth}
    \centering
    \includegraphics[width=\hsize]{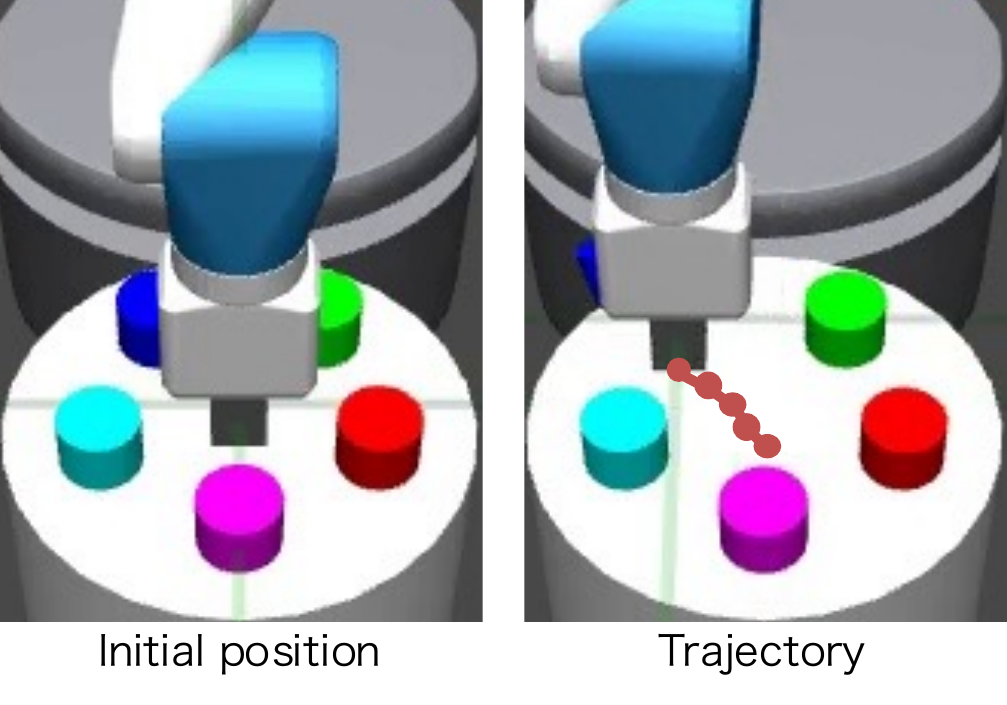}
    \subcaption{Five objects}
    \label{fig:fetch:sweep:trajectory:obj1}
  \end{minipage}
  \hspace{3mm}
  \begin{minipage}[b]{0.3\linewidth}
    \centering
    \includegraphics[width=\hsize]{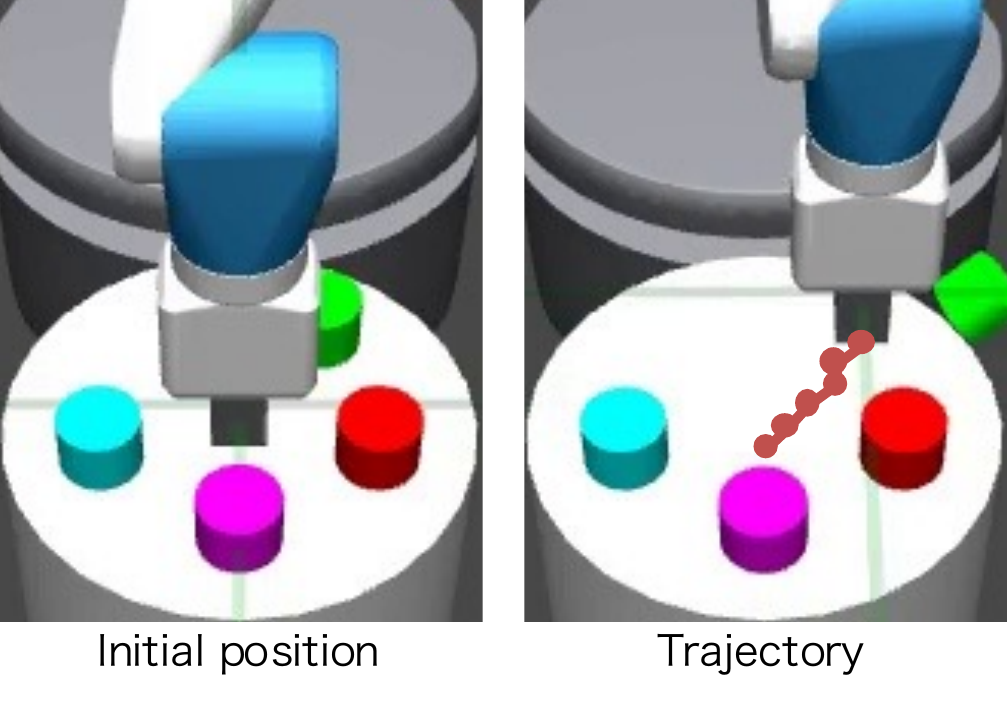}
    \subcaption{Four objects}
    \label{fig:fetch:sweep:trajectory:obj1}
  \end{minipage}
  \hspace{3mm}
  \begin{minipage}[b]{0.3\linewidth}
    \centering
    \includegraphics[width=\hsize]{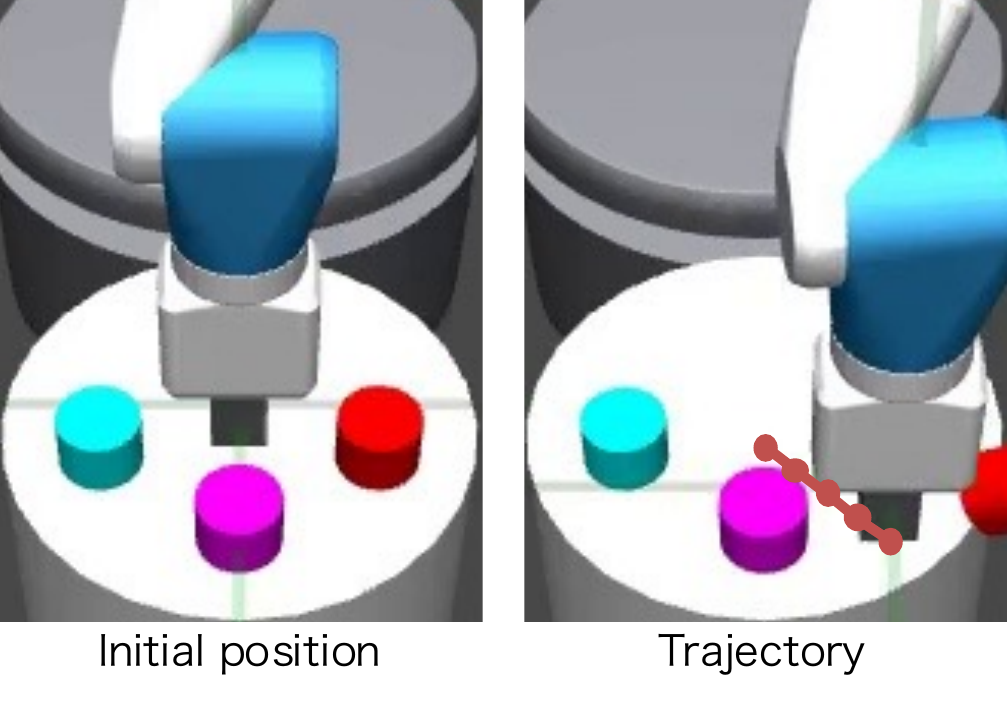}
    \subcaption{Three objects}
    \label{fig:fetch:sweep:trajectory:obj1}
  \end{minipage}\\
  \vspace{3mm}
  \begin{minipage}[b]{0.3\linewidth}
    \centering
    \includegraphics[width=\hsize]{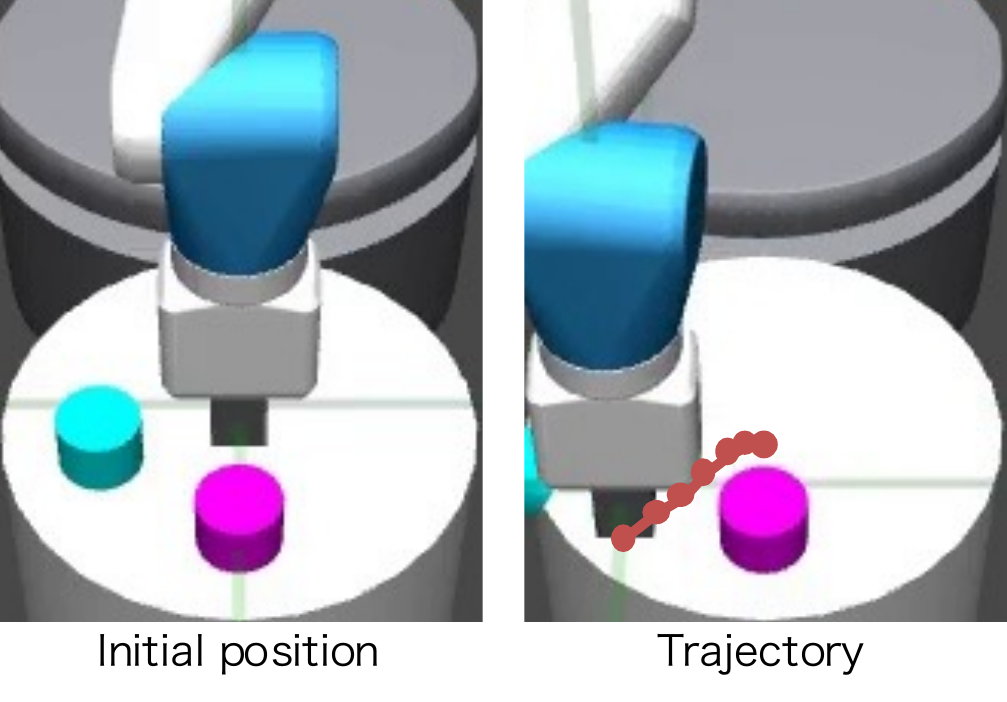}
    \subcaption{Two objects}
    \label{fig:fetch:sweep:trajectory:obj1}
  \end{minipage}
  \hspace{3mm}
  \begin{minipage}[b]{0.3\linewidth}
    \centering
    \includegraphics[width=\hsize]{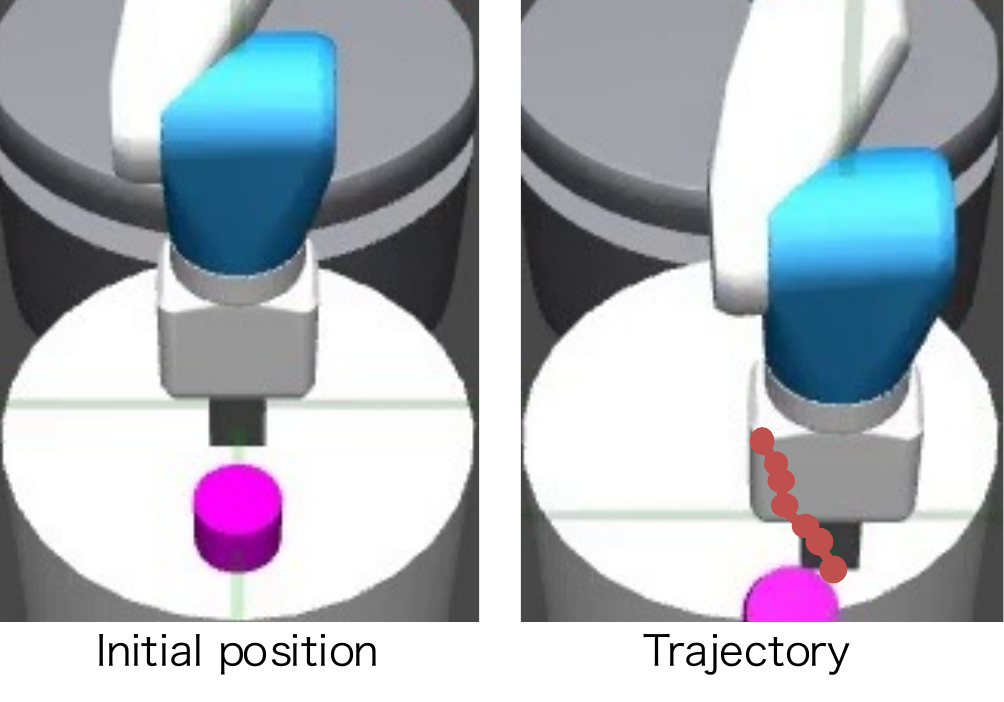}
    \subcaption{One object}
    \label{fig:fetch:sweep:trajectory:obj1}
  \end{minipage}
  \caption{Behaviors of policies learned by multimodal SGP-PS in table-sweeping task}
  \label{fig:fetch:sweep:trajectory}
\end{figure*}
\begin{figure}[!t]
  \centering

  \begin{minipage}[b]{7cm}
    \centering
    \begin{minipage}[b]{3cm}
      \centering
      \includegraphics[width=\hsize]{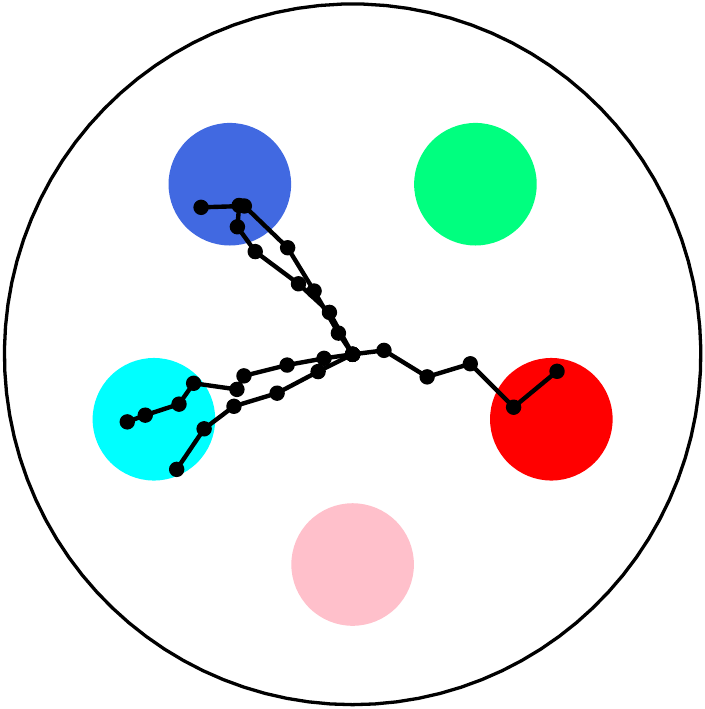}
    \end{minipage}
	\hspace{2mm}
    \begin{minipage}[b]{3cm}
      \centering
      \includegraphics[width=\hsize]{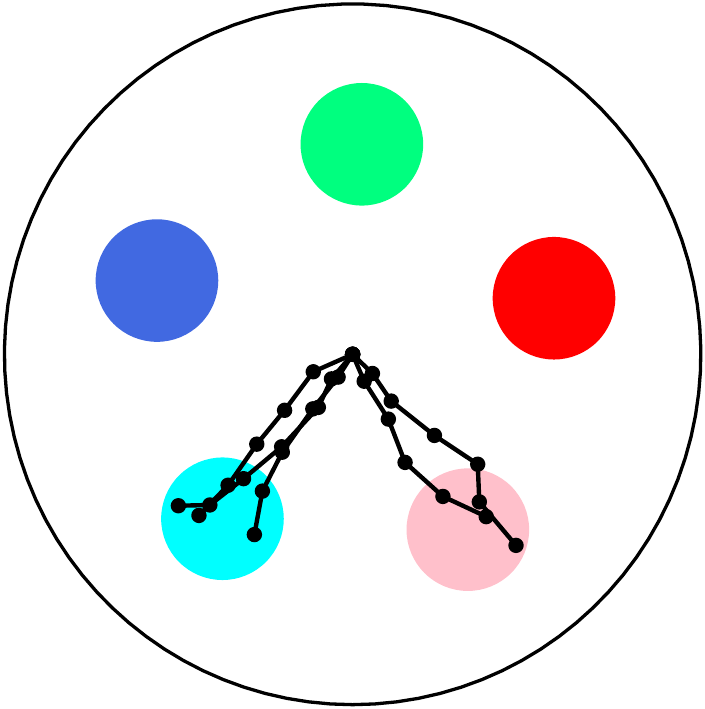}
    \end{minipage}
    \subcaption{Multimodal SGP policy}
  \end{minipage}\\
  \vspace{2mm}
  \begin{minipage}[b]{7cm}
    \centering
    \begin{minipage}[b]{3cm}
      \centering
      \includegraphics[width=\hsize]{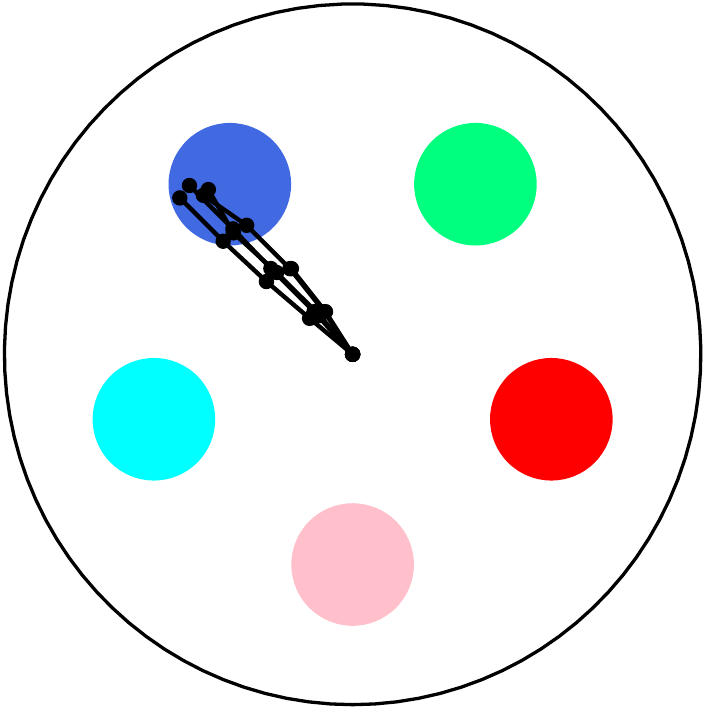}
    \end{minipage}
	\hspace{2mm}
    \begin{minipage}[b]{3cm}
      \centering
      \includegraphics[width=\hsize]{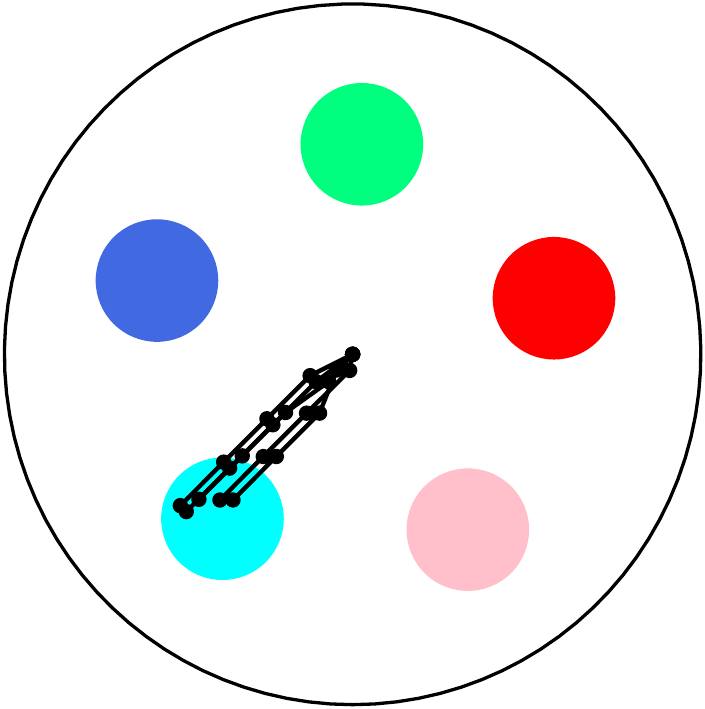}
    \end{minipage}
    \subcaption{Mode-seeking SGP policy}
    \label{fig:fetch:sweep:trajectory:robust}
  \end{minipage}\\
  \vspace{2mm}
  \begin{minipage}[b]{7cm}
    \centering
    \begin{minipage}[b]{3cm}
      \centering
      \includegraphics[width=\hsize]{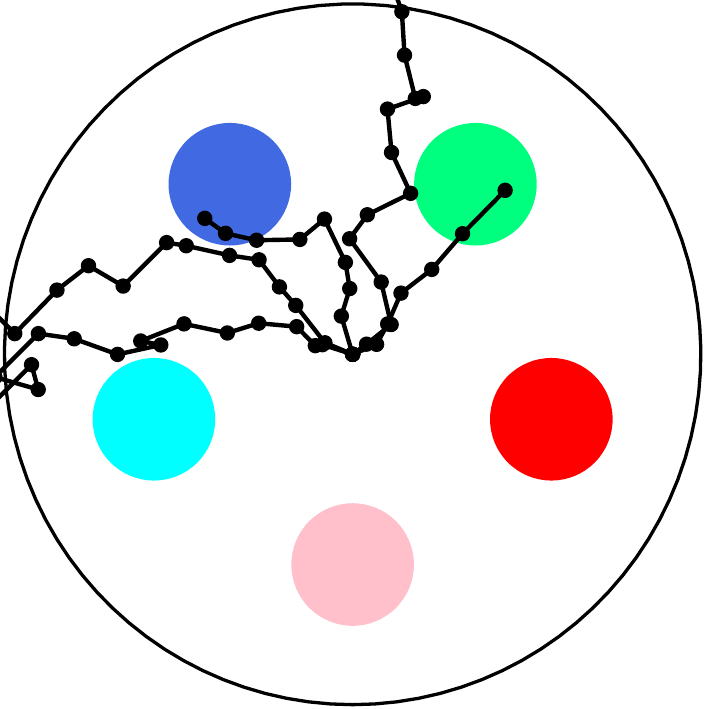}
    \end{minipage}
	\hspace{2mm}
    \begin{minipage}[b]{3cm}
      \centering
      \includegraphics[width=\hsize]{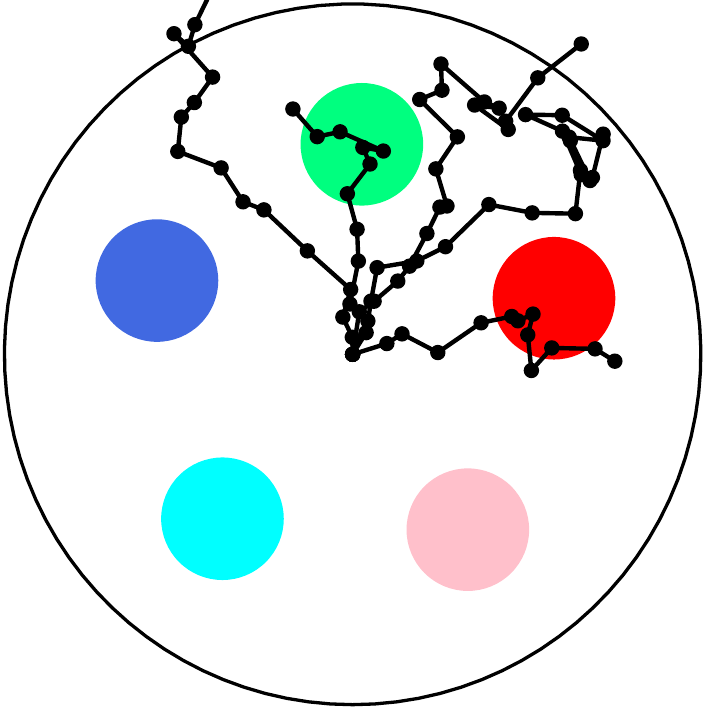}
    \end{minipage}
    \subcaption{Unimodal SGP policy}
    \label{fig:fetch:sweep:trajectory:multi}
  \end{minipage}
  \vspace{-2mm}
  \caption{Five trajectories of the end-effector by learned policy by Multimodal SGP-PS, Mode-seeking SGP-PS, and Unimodal SGP-PS. The left and right figures have different initial positions of five objects. The colored circles indicate the initial positions of five objects on the table. Black lines indicate trajectories by the learned policy.}
  \label{fig:fetch:sweep:trajectory:5objects}
\end{figure}

Figure \ref{fig:fetch:sweep:result} shows the mean and standard deviation of the total reward over ten experiments using multimodal SGP-PS, mode-seeking SGP-PS, unimodal SGP-PS, SAC, and TRPO.
The learning experiment was conducted individually for each number of objects.
In the task with one object, the performance of unimodal SGP-PS is comparable multimodal SGP-PS and mode-seeking SGP-PS.
However, as the number of objects increases, multimodal SGP-PS and mode-seeking SGP-PS outperform unimodal SGP-PS by effectively handling multiple optimal actions.

In the SAC algorithm, although the value function captured multiple optimal actions, the policy employs a unimodal model.
Then it needs to seek one of the optimal actions among multiple ones.
The policy's performance may become unstable because mode-seeking is not stable at the beginning of learning.
Even though TRPO improves the policy monotonically; however, the performance is inadequate.
It needs more data to learn a suitable policy in tasks with multiple optimal actions may be due to neural-network-based policies with many parameters.
This result resembles one that was previously reported \cite{HoofNP17}.

Fig. \ref{fig:fetch:sweep:trajectory} shows the behavior of a policy learned by the multimodal SGP-PS.
For each object, our methods can learn a policy to sweep an object by deciding action multiple steps.

Fig. \ref{fig:fetch:sweep:trajectory:5objects} shows five trajectories generated by the policy learned by multimodal SGP-PS, mode-seeking SGP-PS, and unimodal SGP-PS on the task with five objects.
The multimodal SGP policy learned the multiple optimal actions for sweeping an object in each initial state by the nature of multimodality.
However, it could learn two or three optimal actions out of five due to limited data explored.
The mode-seeking SGP policy learned one of the optimal actions by ignoring others successfully, while the unimodal SGP policy could not learn any optimal actions properly.
We confirmed that each learned policy obtained different optimal actions according to the initial positions of objects.

In summary, all of our simulation results in the table-sweep task suggest the effectiveness of our methods for learning policies with multiple optimal actions.
We confirmed that our methods more effectively learn a policy than SAC and TRPO in a task environment with a small number of samples and multiple optimal actions.

\section{Discussion}
\label{sec:discussion}

We compared multimodal SGP-PS and mode-seeking SGP-PS.
A learned policy by multimodal SGP-PS earned a higher return in simulations.
However, it needs to decide the number of components affecting the performance before learning. 
Therefore, we could select a suitable one from the multimodal SGP-PS and the mode-seeking SGP-PS depending on whether the number of multimodalities of the function is known.

In a mode-seeking SGP-PS, if there is a clear difference in the size of data among multiple optimal actions in the training data, learning tends to proceed faster since it fits the model.
If the multiple optimal actions are observed almost evenly, the mode-seeking SGP-PS may struggle to capture one among them; in that case, the multimodal SGP-PS tends to be faster and more stable than the mode-seeking SGP-PS.
Our experimental results may support the above discussion:
For the table-sweeping task with more than three objects, the multimodal SGP-PS converged learning faster than the mode-seeking SGP-PS.
For the table-sweeping task with one or two objects, the mode-seeking SGP-PS outperformed the multimodal SGP-PS.

Our methods can compute the predictive distribution to determine an action at any state since they employ GP-based policy models.
Our methods explore state-action space by sampling the action using the predictive distribution, i.e., on-policy learning.
In the multimodal SGP-PS, which has multiple components, a softmax function with the uncertainties of all the predictive distributions computes the probability of the component selection.

Although GP provides several advantages for our method, it has limitations in computational complexity.
For tasks that require handling huge amounts of data, our methods require much computational complexity of learning and predictive distribution.
Perhaps our methods are unsuitable for long-horizon tasks.

To apply our methods to complex tasks, we must handle raw images and need more data-efficient algorithms.
Features need to be designed to handle raw images as a state in a non-parametric policy search.
A kernel function that incorporates the structure of convolutional neural networks and deep learning technology specializing in image processing was proposed \cite{garriga-alonsoICLR2019}.
By introducing it into our method, we can learn a policy that uses images as input.
To improve data efficiency, we can introduce a policy update with limited information loss \cite{HoofNP17}.
Them ultimodal SGP-PS needs to decide the number of components since we use overlapping mixtures of GPs inspired by a previous work \cite{OMGP_LAZAhROGREDILLA2012}.
We can also infer the number of components \cite{RossJMLR2013}.

\section{Conclusions}
\label{sec:conclusions}
We proposed a multimodal SGP-PS and a mode-seeking SGP-PS, which are non-parametric policy search methods for a task with multiple optimal actions.
The multimodal SGP-PS employs a multimodal policy prior inspired by OMGP to learn a policy that can capture multiple optimal actions.
The mode-seeking SGP-PS learns a unimodal policy that captures one optimal action by employing an outlier-robust likelihood function.
We derived the updating laws of both methods based on variational Bayesian inference.

To investigate the performance of our methods, we conducted two manipulation tasks: 1) a hand-posture adjustment task and 2) a table-sweeping task.
We confirmed that our methods can learn suitable policies in an environment with multiple optimal actions.

\appendix

\section{Derivation of analytical solutions in multimodal SGP-PS}
\label{appendix:multi}

The analytical solution of variational distribution $q(\bar\bff^{(m)})$ is obtained by solving the following equation:

\begin{align}
  &\log q(\bff^{(m)}) \nonumber\\
  &= \int \left\{ \log p(\tilde\bfa\mid\{\bff^{(m)}\},\bfZ)p(\{\bff^{(m)}\mid\{\bar\bff^{(m)}\})p(\bfZ)\right.\cdot \nonumber \\
  &~~~~~~~\left.p(\{\bar\bff^{(m)}\}) \right\}q(\{\bar\bff^{(i\backslash m)}\})q(\bfZ)\rmd\{\bar\bff^{(i\backslash m)}\}\rmd\bfZ + C
\end{align}
Here $\{\bar\bff^{(i\backslash m)}\}$ indicates all the pseudo outputs without $\bar\bff^{(m)}$.
The following is the analytical solution of $q(\bar\bff^{(m)})$:
\begin{align}
  q(\bar\bff^{(m)}) &= \calN (\bar\bff^{(m)} \mid  \bfmu^{(m)}, \bfSigma^{(m)}), \nonumber \label{eq:multi:update:qf}\\
  \bfmu^{(m)} &= \bfK_{\bar\bfS}^{(m)}\bfQ_L^{(m)^{-1}}\bfK_{\bar\bfS,\bfS}^{(m)}\bfW\bfB^{(m)}\bfW\bfa, \nonumber \\
  \bfSigma^{(m)} &= \bfK_{\bar\bfS}^{(m)}\bfQ_L^{(m)^{-1}}\bfK_{\bar\bfS}^{(m)}, \nonumber \\
  \bfQ^{(m)}_L &= \bfK_{\bar\bfS}^{(m)}+\bfK_{\bar\bfS,\bfS}^{(m)}\bfW\bfB^{(m)}\bfW\bfK_{\bfS,\bar\bfS}^{(m)}, \\
  \bfB^{(m)} &= \mathrm{diag}\left\{ \frac{\hat\bfPi_{nm}}{\lambda^{(m)}_n+\sigma^2} \right\}.
\end{align}

The analytical solution of variational distribution $q(\bfZ)$ is obtained by solving the following equation:
\begin{align}
  &\log q(\bfZ) \nonumber \\
  &= \int \left\{ \log p(\tilde\bfa\mid\{\bff^{(m)}\},\bfZ)p(\{\bar\bff^{(m)}\mid\bff^{(m)})p(\bfZ)\right.\cdot \nonumber \\
  &~~~~~\left.p(\{\bar\bff^{(m)}\}) \right\}q(\{\bar\bff^{(m)}\})\rmd\{\bar\bff^{(m)}\} + C.
\end{align}
The following is the analytical solution of $q(\bfZ)$:
\begin{align}
  q(\bfZ) &= \prod_{n=1,m=1}^{N,M} \hat\bfPi_{nm}^{\bfZ_{nm}}, \label{eq:multi:update:qz} \\
  \hat\bfPi_{nm} &= \bfPi_{nm} \exp(b_{nm}), \nonumber \\
  b_{nm} &= -\frac{1}{2}\log 2\pi(\lambda^{(m)}_n + \sigma^2) \nonumber \\
  & -\frac{\left(\bfW_{nn} a_n-\bfW_{nn}\bfK^{(m)}_{\bfs_n,\bar\bfS}\bfK^{(m)^{-1}}_{\bar\bfS}\bfmu^{(m)} \right)^2}{2(\lambda^{(m)}_n+\sigma^2)}  \nonumber \\
  & -\frac{\left( \bfW_{nn}\bfK_{\bfs_n,\bar\bfS}^{(m)}\bfQ^{(m)^{-1}}_L\bfK_{\bar\bfS,\bfs_n}^{(m)}\bfW_{nn}\right)}{2(\lambda^{(m)}_n+\sigma^2)}. \nonumber
\end{align}

The lower bound of the marginal likelihood is also obtained analytically:
\begin{align}
  &\log J_L'= \nonumber \\
  &\sum_{m=1}^M -\frac{1}{2}\bfa^T\bfW \left(\bfB^{(m)^{-1}} + \bfW\bfK_{\bfS,\bar\bfS}^{(m)}\bfK_{\bar\bfS}^{(m)^{-1}}\bfK_{\bar\bfS,\bfS}^{(m)}\bfW\right)^{-1}\bfW\bfa \nonumber \\
  &+ \sum_{n=1,m=1}^{N,M}\log[\bfR^{(m)}]_{nn} - \mathrm{KL}(q(\bfZ)\mid\mid p(\bfZ)) \nonumber \\
  &-\frac{1}{2} \sum_{n=1,m=1}^{N,M}[\hat\bfPi]_{nm}\log 2\pi (\lambda_n^{(m)}+\sigma^2), \\
  &\bfR^{(m)} = \mathrm{chol} \left( \bfI + \bfB^{(m)^{-1/2}}\bfW\bfK_{\bfS,\bar\bfS}^{(m)}\bfK_{\bar\bfS}^{(m)^{-1}}\bfK_{\bar\bfS,\bfS}^{(m)}\bfW\bfB^{(m)^{-1/2}} \right).
\end{align}

\section{Predictive distribution of multimodal SGP-PS}
Compute the predictive distribution using the variational distribution instead of the true posterior distribution:
\label{appendix:multi:predictive}
\begin{align}
  p(a_*^{(m)}\mid\bfs_*)  &\approx \int p(a_*^{(m)}\mid \bar\bff^{(m)}\bfs_*)q(\bar\bff^{(m)})\rmd\bar\bff^{(m)} \nonumber\\
  &= \calN\left(  a_*^{(m)}\mid  \mu_*^{(m)}, \sigma_*^{(m)} \right), \\
  \mu_*^{(m)} &= \bfK^{(m)}_{\bfs_*,\bar\bfS}\bfQ^{(m)^{-1}}_L\bfK^{(m)}_{\bar\bfS,\bfS}\bfW\bfB^{(m)}\bfW\bfa, \nonumber\\
  \sigma_*^{(m)} &= \bfK_{\bfs_*}^{(m)}-\bfK^{(m)}_{\bfs_*,\bar\bfS}(\bfK_{\bar\bfS}^{(m)^{-1}}-\bfQ_L^{(m)^{-1}})\bfK_{\bar\bfS,\bfs_*}^{(m)}+\sigma^2. \nonumber
\end{align}

\section{Derivation of analytical solutions in mode-seeking SGP-PS}
\label{appendix:robust}

The analytical solution of variational distribution $q(\tau_n)$ is obtained by solving the following equation:
\begin{align}
  &\log q(\tau_n) \nonumber\\
  &= \int \left\{ \log p(\tilde\bfa\mid\bff,\bfT)p(\bff\mid\bar\bff)p(\bfT)p(\bar\bff) \right\} \cdot\nonumber \\
  &~~~~~~~~~q(\{\tau_{i\backslash n}\})q(\bar\bff)\rmd\{\tau_{i\backslash n}\}\rmd\bar\bff + C,
\end{align}
where $\{\tau_{i\backslash n}\}$ indicates all precision variables without $\tau_n$.
The following is the analytical solution of $q(\tau_n)$:
\begin{align}
  q(\tau_n) &= \mathrm{Gam}(\tau_n\mid a_n, b_n), \label{eq:robust:update:qtau}\\
  a_n &= \frac{\nu+1}{2}, \nonumber\\
  b_n &= \frac{\bfW_{nn}^2(\bfa_n-\bfA_n\bfmu)^2}{2}+\frac{\mathrm{tr}(\bfA_n^T\bfW^2_{nn}\bfA_n\bfC)}{2} \nonumber \\
      & +\frac{\bfW_{nn}^2\lambda_n+\nu\sigma^2}{2}, \nonumber \\
  \bfA &= \bfK_{\bfS,\bar\bfS}\bfK_{\bar\bfS}^{-1}. \nonumber
\end{align}

The analytical solution of variational distribution $q(\bar\bff)$ is obtained by solving the following equation:
\begin{align}
  q(\bar\bff) &= \calN(\bar\bff\mid\bfmu,\bfC), \label{eq:robust:update:qf}\\
  \bfmu &= \bfK_{\bar\bfS}\bfSigma^{-1}\bfK_{\bar\bfS,\bfS}\bfW\hat\bfT\bfW\bfa, \nonumber\\
  \bfC &= \bfK_{\bar\bfS}\bfSigma^{-1}\bfK_{\bar\bfS}, \nonumber \\
  \bfSigma &= \bfK_{\bar\bfS,\bfS}\bfW\hat\bfT\bfW\bfK_{\bfS,\bar\bfS} + \bfK_{\bar\bfS}, \nonumber\\
  \hat\bfT &= \mathrm{diag}\{a_n/b_n\}. \nonumber
\end{align}

The lower bound of the marginal likelihood is also obtained analytically:
\begin{align*}
  &\log J'_L = \\
  &-\frac{N}{2}\log 2\pi - \frac{1}{2}\log|\hat\bfT^{-1}| \\
  &- \frac{1}{2} (\bfa-\bfA\bfmu)^T\bfW\hat\bfT\bfW(\bfa-\bfA\bfmu) - \frac{1}{2}\mathrm{tr}(\bfA^T\bfW\hat\bfT\bfW\bfA\bfC) \\
  &- \mathrm{KL}(q(\bar\bff)\mid\mid p(\bar\bff)) - \mathrm{KL}(q(\bfT)\mid\mid p(\bfT))-\frac{1}{2}\mathrm{tr}(\bfW\hat\bfT\bfW\bfLambda). \label{eq:robust:marginal:likelihood}
\end{align*}

\section{Predictive distribution of mode-seeking SGP-PS}
\label{appendix:robust:predictive}
Compute the predictive distribution using the variational distribution instead of the true posterior distribution:
\begin{align}
  p(a_*\mid \bfs_*) &\approx \int p(a_*\mid\bff,\tau)p(\bff\mid\bar\bff)q(\bar\bff)q(\tau)\rmd\bff\rmd\bar\bff\rmd\tau \nonumber\\
  &=\calN(y_*\mid\mu_*,\sigma^2_*),\\
  \mu_* &= \bfK_{\bfs_*,\bar\bfS}\bfSigma^{-1}\bfK_{\bar\bfS,\bfS}\bfW\hat\bfT\bfW\bfa, \nonumber\\
  \sigma^2_* &= \bfK_* - \bfK_{\bfs_*,\bar\bfS}\bfK_{\bar\bfS}^{-1}\bfK_{\bar\bfS,\bfs_*} + \bfK_{\bfs_*,\bar\bfS}\bfSigma^{-1}\bfK_{\bar\bfS,\bfs*} + \sigma^2. \nonumber
\end{align}

\section{Hyperparameters of SAC and TRPO in the table-sweeping task}
This section describes the hyperparameters of SAC and TRPO.
\begin{table}[!h]
  \caption{SAC hyperparameters}
  \vspace{-3mm}
  \label{table:sac:params}
  \centering
  \begin{tabular}{|c|c|} \hline
    Parameter & Value \\ \hline\hline
    optimizer & Adam \\
    learning rate & $3\times 10^{-4}$ \\
    discount ($\gamma$) & 0.99 \\
    replay buffer size & $10^6$ \\
    number of hidden layers & 2 \\
    number of hidden units per layer & 128 \\
    number of sample per minibatch & 256 \\ 
    nonlinearity & ReLU \\ \hline
  \end{tabular}
\end{table}
\begin{table}[!h]
  \caption{TRPO hyperparameters}
  \vspace{-3mm}
  \label{table:trpo:params}
  \centering
  \begin{tabular}{|c|c|} \hline
    Parameter & Value \\ \hline\hline
    optimizer & Adam \\ 
    learning rate & $1\times 10^{-3}$ \\ 
    discount ($\gamma$) & 0.995 \\ 
    Stepsize ($\mathrm{KL}$) & 0.01 \\ 
    replay buffer size & $10^6$ \\
    number of hidden layers & 2 \\
    number of hidden units per layer & 64 \\
    number of sample per minibatch & 256 \\ 
    nonlinearity & Tanh \\ \hline
  \end{tabular}
\end{table}

\bibliographystyle{elsarticle-num}
\bibliography{journal_ref}

\end{document}